\documentclass[conference,compsoc]{IEEEtran}

\usepackage[utf8]{inputenc} 
\usepackage[T1]{fontenc}    
\usepackage{hyperref}       
\usepackage{url}            
\usepackage{booktabs}       
\usepackage{amsfonts}       
\usepackage{nicefrac}       
\usepackage{microtype}      
\usepackage{xcolor}         
\usepackage{graphicx}
\usepackage[linesnumbered,ruled,vlined]{algorithm2e}
\usepackage{multirow}
\usepackage{amsmath}
\usepackage{subfigure}
\usepackage{xcolor}
\usepackage{cite}
\usepackage{float}

\begin{document}

\title{Is Diffusion Model Safe? Severe Data Leakage via Gradient-Guided Diffusion Model}

\author{%
Jiayang Meng$^{1*}$ \quad Tao Huang$^{2*}$ \quad Hong Chen$^{1\textsuperscript{\textdagger}}$ \quad Cuiping Li$^1$\\ 
$^1$ School of Information, Renmin University of China\\
$^2$ School of Computer Science and Big Data, Minjiang University\\
\texttt{\{jiayangmeng,chong,licuiping\}@ruc.edu.cn}\\
\texttt{\{huang-tao\}@mju.edu.cn}
}


\maketitle

\begin{abstract}
Gradient leakage has been identified as a potential source of privacy breaches in modern image processing systems, where the adversary can completely reconstruct the training images from leaked gradients. However, existing methods are restricted to reconstructing low-resolution images where data leakage risks of image processing systems are not sufficiently explored. In this paper, by exploiting diffusion models, we propose an innovative gradient-guided fine-tuning method and introduce a new reconstruction attack that is capable of stealing private, high-resolution images from image processing systems through leaked gradients where severe data leakage encounters. Our attack method is easy to implement and requires little prior knowledge. The experimental results indicate that current reconstruction attacks can steal images only up to a resolution of $128 \times 128$ pixels, while our attack method can successfully recover and steal images with  resolutions up to $512 \times 512$ pixels. Our attack method significantly outperforms the SOTA attack baselines in terms of both pixel-wise accuracy and time efficiency of image reconstruction. Furthermore, our attack can render differential privacy ineffective to some extent.
\end{abstract}


%
\IEEEpeerreviewmaketitle

\section{Introduction}

In the domain of image processing, the training images are often private and carry private information about the providers. For instance, if someone's face image is leaked, the leaked image can be used to unlock personal devices. Distributed or federated training offers protection for locally sensitive data to some extent. In those paradigms, the exchange of gradients, rather than the exchange of private data, is executed in parallel across multiple participants. This process can be facilitated by either a parameter server approach or an all-reduce strategy, as detailed in \cite{b1,b2} and \cite{b3,b4} respectively. The inherent nature of distributed or federated learning that allows each client to retain its training data locally and share only gradients is particularly beneficial when dealing with sensitive information, as it enables the pooling of data from various sources without centralizing it, as highlighted in \cite{b5,b6}. A practical application is observed in the healthcare sector, where multiple hospitals collaboratively train a model without sharing sensitive patient data, as demonstrated in \cite{b7,b8}.

Recent studies on the risks associated with reconstructing private training images in distributed or federated scenarios reveal that adversaries can exploit gradients to create images closely resembling private images \cite{b9,b10,b11} or even to replicate the private images themselves \cite{b12,b13}. A leading method in this field, \textit{Deep Leakage from Gradients} (DLG) \cite{b12,b13}, stands out for only using gradients to achieve pixel-wise accurate image reconstruction, contrasting with traditional methods that require additional information and typically yield only partial or synthetic representations of the target image. While there are various advanced methods \cite{b12,b13,ig,gi,gias,ggl,gifd} focusing on image reconstruction, the privacy risks in distributed or federated learning systems remain under-explored, particularly concerning the reconstruction of high-resolution images. Stealing high-resolution images is necessary, as there are many situations in which high-resolution images are needed while low-resolution ones do not work. For instance, in fields like radiology, high-resolution images can make the difference between missing and identifying critical patient conditions. Detailed images help in diagnosing diseases, planning treatments, and monitoring patient progress. These images are costly to access and always contain private information. Attackers are highly motivated to illicitly acquire these high-resolution images due to commercial interests or other reasons. However, existing state-of-the-art (SOTA) methods struggle significantly on this issue. For example, DLG's attempts at reconstructing high-resolution images tend to be unsuccessful, resulting in distorted visuals and excessive noise. Developing effective techniques to reconstruct high-resolution images is essential to identify potential privacy vulnerabilities in image processing systems and to advance research on secure image utilization during training phases.

Reconstructing high-resolution images poses a significant challenge due to the intricate process of establishing the mappings between leaked gradients and the original images. Diffusion models, including Denoising Diffusion Probabilistic Models (DDPM) \cite{b14,b15}, Denoising Diffusion Implicit Models (DDIM) \cite{b16}, and so on, have become popular in image generation tasks. DDIM, a variant of DDPM, stands out for its excellent ability to generate high-resolution images stably. A key attribute of diffusion models lies in their versatility: the image generation process can be conceptualized as the mapping between latent space and original image space and can be guided by a score function \cite{b18}, text descriptions \cite{b19,b20,b21,b22}, or reference images \cite{b19}, thereby enhancing their generative capability.

However, existing guidance to diffusion models mentioned above can not be directly utilized in reconstruction attacks since attackers can only touch gradients in image processing systems. Our research explores the fine-tuning method via only gradients and equips diffusion models with the capability of stealing high-resolution images using leaked gradients. In detail, We propose a novel and time efficient method where a pre-trained diffusion model is fine-tuned with the guidance of leaked gradients, enabling the reconstruction of high-resolution images via the fine-tuned diffusion model. And we are pleasantly surprised to discover that our method appears to exhibit resilience to differential privacy. This study aims to underscore the privacy vulnerabilities associated with gradient exchanges in machine learning systems and emphasizes the pressing need for innovative solutions to address these security concerns.

\section{Related Work}

\subsection{Gradient Leakage and Image Reconstruction}\label{subsection2.1}

In deep learning, model training is executed in parallel across different nodes sometimes, and synchronization occurs through gradient exchange, for example, federated learning \cite{b6}. However, it is indeed possible to extract private training data from the shared gradients \cite{b9,b10,b11,b12,b13}. Namely, the adversary can recover the training inputs from shared gradients, questioning the safety of gradient sharing in preserving data privacy. An effective method is DLG (Deep Leakage from Gradients) \cite{b12,b13}, where the attacker optimizes a dummy input image to minimize the difference between its gradient in the model and the leaked gradient. The optimization can be formulated as:
\begin{equation}
    \begin{split}
    \min _{x^{\prime}}||\nabla F\left( x^{\prime},W \right) -\nabla F\left( x,W \right) ||^2
    \end{split}
\label{eq1}
\end{equation}
where $x^{\prime}$ is the dummy input initialized randomly, $\nabla F\left( x^{\prime}, W \right)$ is the gradient of the loss with respect to $x^{\prime}$ for the current model weights $W$, and $\nabla F\left( x, W \right)$ is the leaked gradient that the attacker obtains. $F$ is the loss function.

In the realm of reconstructing training inputs with shared gradients, various advanced methodologies \cite{ig,gi,gias,ggl,gifd} have emerged. \cite{ig} achieves efficient image recovery through a meticulously designed loss function. \cite{gi,gias,ggl} propose the utilization of Generative Adversarial Networks (GANs) acting as prior information, which offers an effective approximation of natural image spaces, significantly enhancing attack methodologies. \cite{gifd} exhibits outstanding generalization capabilities in practical settings, underscoring its adaptability and wider applicability across various scenarios. Although these methods have made progress in the various focal points they emphasize, no specific research has been conducted yet with the aim of enhancing the resolution of image reconstruction.

\subsection{Diffusion Models}\label{subsection2.2}

Diffusion models \cite{b14,b15,b16,b17}, inspired by the physical process of diffusion, which describes the movement of particles from regions of higher concentration to lower concentration, are a class of generative models in machine learning that have gained significant attention due to their ability to produce high-quality, diverse samples. Diffusion models work by gradually adding noise to data over a series of steps, transforming the data into a pure noise distribution. This process is then reversed to generate new data samples from noise. The two key processes involved are the forward diffusion process and the reverse generation process. 

Formally, the forward process is a Markov chain that incrementally adds Gaussian noise to the data at each step. If $x_0$ represents the original data, the process generates a sequence $\{ x_1, x_2, \ldots, x_T \}$ where each $x_t$ is a noisier version of $x_{t-1}$, and $x_T$ is indistinguishable from pure noise. This can be mathematically represented as \cite{b14,b15}:
\begin{equation}
x_t=\sqrt{\alpha_t}x_{0}+\sqrt{1-\alpha_t}\epsilon _t, \epsilon_t \sim \mathcal{N}\left( 0,I \right)
\label{forward}
\end{equation}
where $\alpha_t: = \prod_{s=1}^t \left(1 - \beta_s \right)$, $\{\beta_t \}_{t=1}^{T}$ are fixed variance schedules and $\epsilon_t$ is Gaussian noise. Both DDPM and DDIM have the same forward diffusion process as Eq.(\ref{forward}).

The reverse process is a denoising process. It involves learning a parameterized model $p_{\theta}\left( x_{t-1}|x_t \right)$ to reverse the forward steps. It is modeled by a neural network trained to predict the noise $\epsilon_t$ that is added at each step in the forward process. The sampling methods of different diffusion models differ. In DDPM, the reverse process is a stochastic Markovian process, which is formulated as \cite{b14,b15}:
\begin{equation}
x_{t-1}=\frac{1}{\sqrt{1- \beta_t}}\left( x_t-\frac{\beta_t}{\sqrt{1-\alpha_{t}}}\epsilon _{\theta}\left( x_t,t \right) \right) + \sigma_t z
\label{eq2}
\end{equation}
where $\epsilon _{\theta}\left( x_t,t \right)$ is a prediction of $\epsilon_t$, and $z \sim \mathcal{N}\left( 0,I \right)$. Meanwhile, DDIM \cite{b16} is proposed as an alternative non-Markovian denoising process that has a distinct sampling process as follow \cite{b16}:
\begin{equation}
    \begin{split}
    x_{t-1}=\sqrt{\alpha_{t-1}} f_{\theta}\left(x_{t}, t\right)+\sqrt{1-\alpha_{t-1}-\sigma_{t}^{2}} \epsilon_{\theta}\left(x_{t}, t\right)+\sigma_{t}^{2} z{\scriptsize } 
    \end{split}
\label{DDIM_1}
\end{equation}
where, $z \sim \mathcal{N}\left( 0,I \right)$, $\sigma_{t}=0$ and $f_{\theta}\left(x_{t}, t\right)$ is the prediction of $x_{0}$ at $t$ given $x_{t}$ and $\epsilon_{\theta}\left(x_{t}, t\right)$:
\begin{equation}
f_{\theta}\left(x_{t}, t\right):=\frac{x_{t}-\sqrt{1-\alpha_{t}} \epsilon_{\theta}\left(x_{t}, t\right)}{\sqrt{\alpha_{t}}}
\label{DDIM_2}
\end{equation}

Eq.(\ref{DDIM_2}) is derived from Eq.(\ref{forward}). In DDIM, $\sigma_{t}=0$, so its reverse diffusion process is deterministic, resulting in its ability to generate images stably, which suits our motivation and is used in our attack.

Training diffusion models involves optimizing the parameters $\theta$ of the reverse process, which tries to minimize the difference between the noise added in the forward step $t$ and the predicted noise in the reverse step $t$, in simple terms, makes the generated image similar to the input one. The optimization objective can be expressed as \cite{b14,b15}:
\begin{equation}
\min _{\theta}\mathbb{E} _{t,x_0,\epsilon _t}\left[ ||\epsilon _t-\epsilon _{\theta}\left( x_t,t \right) ||^2 \right]
\label{eq3}
\end{equation}

\section{Gradient as Embedding: Stealing Images via Gradient-Guided Diffusion Model}

In this study, we analyze an adversarial interaction scenario with a deep learning model, denoted as $F(W)$, wherein $W$ represents the set of weights characterizing the model. Stochastic Gradient Descent (SGD) is commonly used to update weights $W$ of the model $F(W)$. Given a data $x$ sampled from the training set, its gradient is calculated as:
\begin{equation}
\nabla F\left( x,W \right) =\frac{\partial F\left( x,W \right)}{\partial W}      
\label{eq7}
\end{equation}

\begin{figure*}
  \centering
  \includegraphics[scale=0.08]{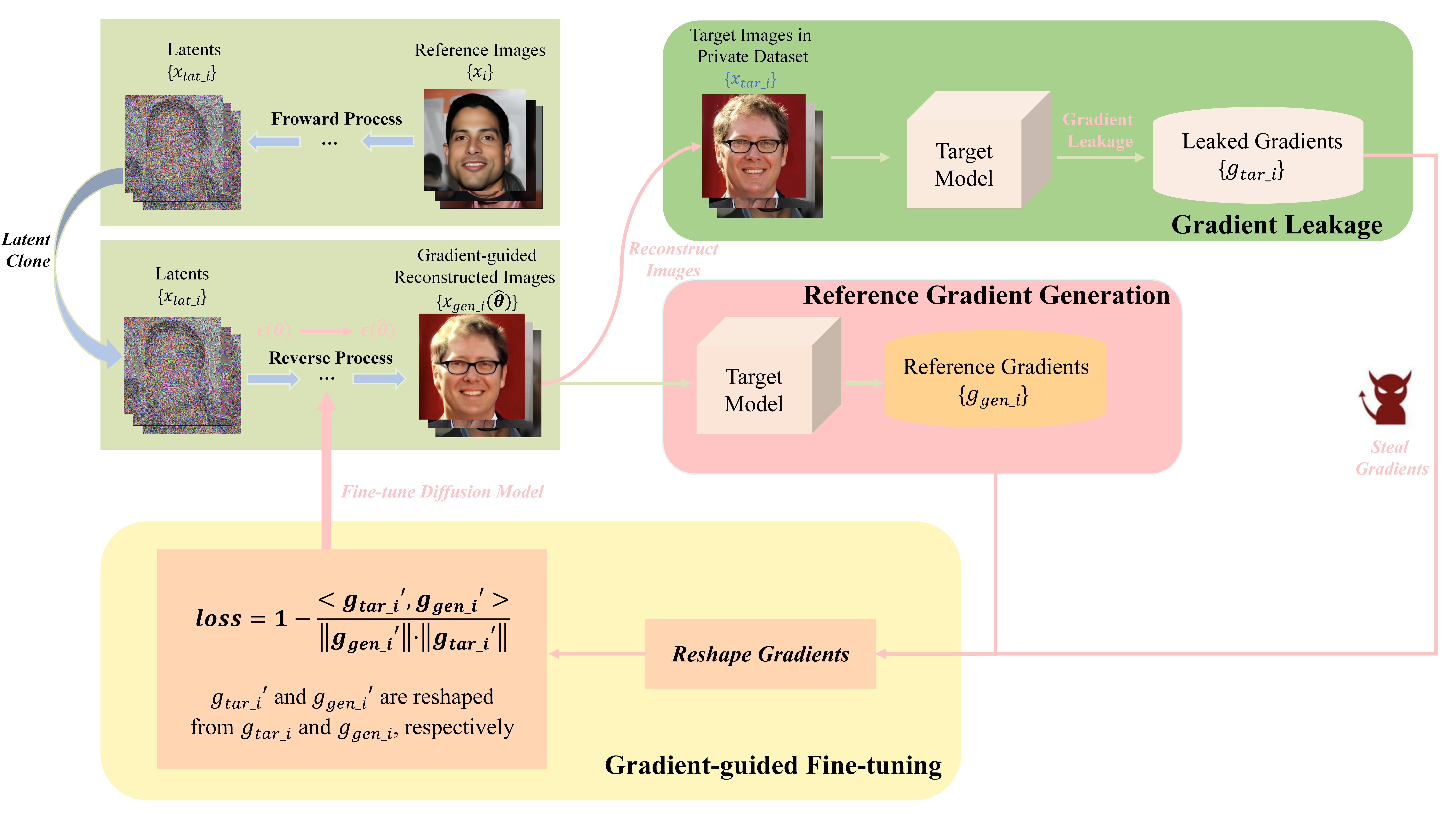}
  \caption{Image reconstruction via gradient-guided diffusion model.}
  \label{fig1_main}
\end{figure*}

The gradient may be conceptualized as a latent representation of a data sample, with the machine-learning model functioning analogously to an encoder. Consequently, the gradient encapsulates sample-specific information, which is instrumental in gradient-guided fine-tuning. We hypothesize that an adversary is capable of exfiltrating the gradient, $g_{\text{tar}} = \nabla F\left( x_{\text{tar}}, W \right)$, computed with a particular data sample, $x_{\text{tar}}$, from the private training dataset of the machine-learning model. The acquisition of this gradient highlights the potential security vulnerability within the training data's protection framework. Our method for image reconstruction leveraging the outstanding image generation capability of diffusion models is depicted in Figure \ref{fig1_main}.

\subsection{Gradient-Guided Fine-tuning}\label{Section3.1}

The adversary is presumed to have access to a public pre-trained diffusion model, represented as $\epsilon(\theta)$, where $\theta$ denotes the parameters of this diffusion model. Utilizing this model, the adversary is capable of synthesizing an image $x_{\text{gen}}(\theta)$, which has the diffusion model parameters $\theta$ and is subsequently input into the target attacked model. This process results in the generation of the corresponding gradient, $g_{\text{gen}} = \nabla F\left( x_{\text{gen}}(\theta), W \right)$, providing further insights into the model's behavior.

The adversarial strategy progresses to the refinement of the diffusion model, resulting in a fine-tuned diffusion model $\epsilon(\hat{\theta})$. This refinement is achieved by optimizing the initial parameters $\theta$ of the pre-trained diffusion model, leading to the optimized parameter set $\hat{\theta}$. This optimization process is expressed as follows:

{\small
\begin{equation}
    \begin{split}
    \hat{\theta}=\underset{\theta}{\mathrm{argmin}} \left( 1-\frac{<\nabla F\left(x_{gen}\left(\theta \right),W \right) ,\nabla F\left(x_{tar},W \right)>}{||\nabla F\left(x_{gen}\left(\theta \right),W \right)||\cdot||\nabla F\left(x_{tar},W \right)||}\right) 
    \end{split}
\label{eq5}
\end{equation}
}

In the context of machine learning, particularly in the scenario where a model is tasked with classification, it is feasible for an adversary to infer certain characteristics of the private dataset used to train the model. We explore the implications of such inference. Specifically, for example, in cases where the target model is designed to classify gender, an adversary can reasonably hypothesize that the underlying dataset comprises images of individuals identified as male and female. The adversary, aiming to reconstruct images from this private dataset, may opt to employ a diffusion model pre-trained on a dataset containing similar demographic representations (i.e., images of men and women), which is not a mandatory requirement. Then, the gradient guidance may work more effectively. This process involves fine-tuning the outputs of the model to more closely align with the specific attributes and distribution of the target dataset, thereby increasing the likelihood of accurately reconstructing images from the private dataset. This exploration underscores the potential vulnerabilities in machine learning models, particularly in the context of maintaining the confidentiality of private datasets.
\subsection{Pipelines in Stealing Images}
The image reconstruction pipelines consist of:

(1) \textit{Stealing Gradients}: This initial stage involves the adversary's extraction of gradients with respect to images from the target attacked model's private training dataset. This step is critical as it lays the groundwork for subsequent stages by providing essential gradient information. The adversary can pretend to be a participant in federated learning or distributed learning and steals the gradients when gradient transmission happens.

(2) \textit{Selection of a Diffusion Model for Fine-tuning}: The adversary can conduct a thorough examination of the universal distribution characteristics inherent in the target model. This analysis helps to determine the most suitable pre-trained diffusion model for further fine-tuning. Specifically, as mentioned in Subsection \ref{subsection2.2}, we use DDIM in our attack. However, there are no specific mandatory requirements for the pre-trained DDIM, as long as the resolution of the input of the model is the same as that of the attacked image. The chosen model is then prepared for gradient-guided fine-tuning by the gradients previously obtained.  

(3) \textit{Fine-tuning Diffusion Model and Reconstructing Images}: In this phase, the adversary samples a set of reference images from the public training dataset of the pre-trained diffusion model, the selection of which has no impact on reconstruction results. The reference images are fed into the forward process of the diffusion model to obtain and save the corresponding latents. Then, each latent is used to reconstruct a target image, one at a time, to improve the reconstruction quality. The batch size of SOTA methods also equals 1. A larger batch size degrades reconstruction performance and can not improve time efficiency. Regarding the reconstructing process, each latent is used to generate an image through the reverse process of the diffusion model with the parameters $\theta$. The image is then input into the target model to acquire the corresponding reference gradient. The diffusion model undergoes fine-tuning, governed by Eq.(\ref{eq5}), a process that adjusts the model parameters to enhance performance in the image reconstruction task. It is actually necessary to fine-tune a unique diffusion model for each leaked gradient, as the purpose of our fine-tuning is confining the outputs of the fine-tuned model to be the reconstructed private images, that is, we aim for the model's outputs to overfit the target images. Upon fine-tuning the diffusion model for each target gradient, the adversary employs this fine-tuned model to generate image, which constitutes the reconstructed image. This stage marks the culmination of the pipeline, where the fine-tuned model's efficacy in image reconstruction is realized.

\subsection{Algorithm Details}

The algorithm of our method is presented as Algorithm \ref{alg1}. Initially, reference images are fed into the diffusion model and undergo a forward process to transition into latent representations, which are saved and subsequently employed in the reverse process aimed at reconstructing images. During the preliminary phase, the reconstructed images may exhibit significant deviation from the target images, because as mentioned in Subsection \ref{subsection2.2} the pre-trained diffusion model should generate an image that is similar to the input. To refine the reconstruction quality, the proposed method incorporates a unique gradient-guided fine-tuning mechanism. The reconstructed images are processed through the attacked model to extract reference gradients. These gradients and the leaked gradients are then reshaped to vectors and compare the similarity between them using cosine similarity expressed in Eq.(\ref{eq8}) as the fine-tuning loss function for the $i-th$ target image. After finishing fine-tuning, the diffusion model demonstrates enhanced capability in generating images that are similar to the target images within the private dataset. 
\begin{equation}
\mathcal{L}\left(\hat{\theta}\right) =  1-\frac{<\nabla F\left( x_{gen}\left( \hat{\theta} \right) ,W \right) ,\nabla F\left( x_{i},W \right) >}{||\nabla F\left( x_{gen}\left( \hat{\theta} \right) ,W \right) ||\cdot ||\nabla F\left( x_{i},W \right)||}    
\label{eq8}
\end{equation}

\begin{algorithm*}[tb]
    \caption{Image reconstruction via gradient-guided diffusion model.}
    \label{alg1}
    \KwData{A Pre-trained Diffusion Model $\epsilon(\theta)$; The Public Training Set $D_{\text{diff}}$ of $\epsilon(\theta)$; A Differentiable Machine Learning Model $F\left( x,W \right)$; Leaked Gradients $\{ \nabla F\left( x_i,W \right)\}_{i=1}^I$ calculated by training images $\{x_1,\ldots,x_I\}$; Fine-tuning Iteration $T$; Initial Learning Rate $\eta$; Decay Rate of Learning Rate $\gamma$.}
    \KwResult{Fine-tuned Diffusion Model $\{\hat{\theta_1}',\ldots,\hat{\theta_I}'\}$; Reconstructed Images $\{x_1',\ldots,x_I'\}$.}
    \While{Not Finished}{
        \tcp*[h]{Step 1: Gradient-Guided Diffusion Model Fine-tuning.}\\
        Sample a reference image $x_{\text{ref}} \in D_{\text{diff}}$ \\
        Generate an image $x_{gen}\left( \theta \right) \sim \epsilon(\theta, x_{\text{ref}})$ \\
        Calculate gradient $\{ \nabla F\left( x_{gen}\left( \theta \right),W \right)\}$ \\
        \For{$i \in \{1,2,\cdots,I\}$}{
            \For{$t=1$ \KwTo $T$}{
                \If{$t=1$}{
                    $lr \leftarrow \eta$
                }
                \Else{
                    $lr \leftarrow \gamma \cdot lr$
                }
                Update $\hat{\theta}_{t} \longleftarrow \hat{\theta}_{t-1} - lr  \frac{\partial \mathcal{L}\left(\hat{\theta}_{t-1}\right)}{\partial \theta}$ where $\hat{\theta}_{0} \leftarrow \theta$
            }
            Save $\hat{\theta}_{T}$ as $\hat{\theta}_{i}'$
        }
        \tcp*[h]{Step 2: Image Reconstruction via Fine-tuned Diffusion Model.}\\
        \For{$i \in \{1,2,\cdots,I\}$}{
            Reconstruct images via $\epsilon(\hat{\theta}_{i}')$: $x_{i}' \leftarrow x_{gen}\left( \hat{\theta}_{i}' \right) \sim \epsilon(\hat{\theta}_{i}', x_{\text{ref}})$
        }
    }
    \KwRet{Reconstructed Images $\{x_1',\ldots,x_I'\}$.}
\end{algorithm*}

\section{Experiments}

\subsection{Experimental Setup}\label{subsection.4.1}

\textbf{Machine Configuration.} All experiments are run over a GPU machine with one Intel(R) Xeon(R) Gold 5218R CPU @ 2.10GHz with 251 GB DRAM and 4 NVIDIA RTX A6000 GPUs (each GPU has 48 GB DRAM). The operating system is Linux. We implement our attack with Pytorch 2.0.1.

\textbf{Basic Settings.} We choose one low-resolution dataset and four high-resolution datasets as private datasets: CIFAR\_10 \cite{b24}($32 \times 32$), CelebA-HQ \cite{b25}($256 \times 256$), LSUN\_Bedroom \cite{b26}($256 \times 256$), LSUN\_Chruch \cite{b26}($256 \times 256$) and ImageNet \cite{b27}($512 \times 512$). The pre-trained diffusion model is DDIM \cite{b16}. And we choose the same target attacked model as \cite{b12,b13}. Other hyperparameter settings, like the return step $t_0$, the pair of the forward and reverse steps $(S_{\text{for}}, S_{\text{gen}})$ and so on, are shown in Table \ref{settings}.

\begin{table*}[htp!]
   \caption{Hyperparameter settings.}
   \label{settings}
   \centering
   \scalebox{1}{
       \begin{tabular}{c|c|c|c|c|c|c}
          \hline
          \toprule
          \textbf{\multirow{2}{*}{Dataset}}       & \textbf{\multirow{2}{*}{Size of Images}}    & \textbf{Optimizer of} & \textbf{Initial} & \textbf{Decay Rate of} & \textbf{Return Step}    & \textbf{\multirow{2}{*}{$(S_{\text{for}},S_{\text{gen}})$}} \\ 
               &     & \textbf{Fine-tuning Process} & \textbf{Learning Rate} & \textbf{Learning Rate} & \textbf{$t_0$}    &  \\ 
          \midrule
          \textbf{CIFAR\_10}       & $32 \times 32$   & \multirow{5}{*}{Adam Optimizer}  & 6e-5                  & 0.9990     & \multirow{5}{*}{500} & \multirow{4}{*}{(40,6)}           \\
          \textbf{CelebA-HQ}     & $256 \times 256$ &                                  & 2e-5                  & 0.9800     &                      &                                   \\
          \textbf{LUSN\_Bedroom} & $256 \times 256$ &                                  & 1e-4                  & 0.9900     &                      &                                   \\
          \textbf{LUSN\_Church}  & $256 \times 256$ &                                  & 9.5e-5                & 0.9990     &                      &                                   \\ \cline{7-7} 
          \textbf{ImageNet}      & $512 \times 512$ &                                  & 8e-5                  & 0.9997     &                      & (40,2)                            \\ 
          \bottomrule
        \end{tabular}
    }
\end{table*}

\textbf{Evaluation Metrics.} We measure the quality of reconstructed images by adopting  four metrics commonly used in computer vision to compare the similarity between the image and the target image, which are \textit{MSE (Mean Square Error)}, \textit{SSIM (Structural Similarity Index)}, \textit{PSNR (Peak Signal-to-Noise Ratio)}, and \textit{LPIPS (Learned Perceptual Image Patch Similarity)}.

\textit{MSE.} MSE is used to quantify the difference between a target image and the relevant reconstructed image. A lower MSE value indicates that the reconstructed image is closer to the target image, implying better reconstruction quality. 

\textit{SSIM.} SSIM is also used for measuring the similarity between two images, which is considered to be more perceptually relevant than traditional methods like MSE since it incorporates changes in structural information, luminance, and contrast. SSIM ranges from -1 to 1. A value of 1 indicates perfect similarity, and it is desirable to have a higher SSIM when evaluating reconstruction quality.

\textit{PSNR.} PSNR is a popular metric used the assessment of image reconstruction quality. It measures the ratio between the maximum possible power of a signal (in our experiments, the target image) and the power of distorting noise that affects its representation (in our experiments, the reconstructed image). A higher PSNR value implies that the reconstruction is of higher quality. In contrast, a lower PSNR means more error or noise.

\textit{LPIPS \cite{b28}.} Unlike SSIM and MSE, which are handcrafted metrics, LPIPS leverages deep learning to more closely align with human perceptual judgments. LPIPS measures perceptual similarity by extracting deep features from images using a pre-trained convolutional neural network (CNN) and calculating the distance (typically Euclidean) among these feature representations. A lower LPIPS score denotes higher similarity and better reconstruction quality.

\textbf{Qualitative Comparison.} In our experiments, we conduct a qualitative assessment of manipulation performance using gradient-based reconstruction techniques. Due to the impressive efficacy of Deep Leakage from Gradients (DLG) in reconstructing images, we primarily select it as our benchmark for a detailed comparison with our method. We evaluate our proposed method against DLG across two distinct image resolutions: low-resolution ($32 \times 32$ pixels) and high-resolution ($256 \times 256$ pixels and $512 \times 512$ pixels). In addition, to thoroughly show the performance of our method in high-resolution image reconstruction, we compare our method with other baselines mentioned in Subsection \ref{subsection2.1}. Besides assessing reconstruction quality, we also focus on the time efficiency and the resilience to differential privacy protection of the reconstruction process.

\subsection{Experiment Results}

\subsubsection{Comparison with Stealing Capability of DLG}

We perform a comprehensive analysis to compare the results of our method with those of DLG, focusing on the following two aspects.

\begin{figure*}[htp!]
    \centering
    \vspace{-2mm}
    \subfigure[Deep Leakage from Gradients (DLG)]{
		\includegraphics[width=0.48\textwidth]{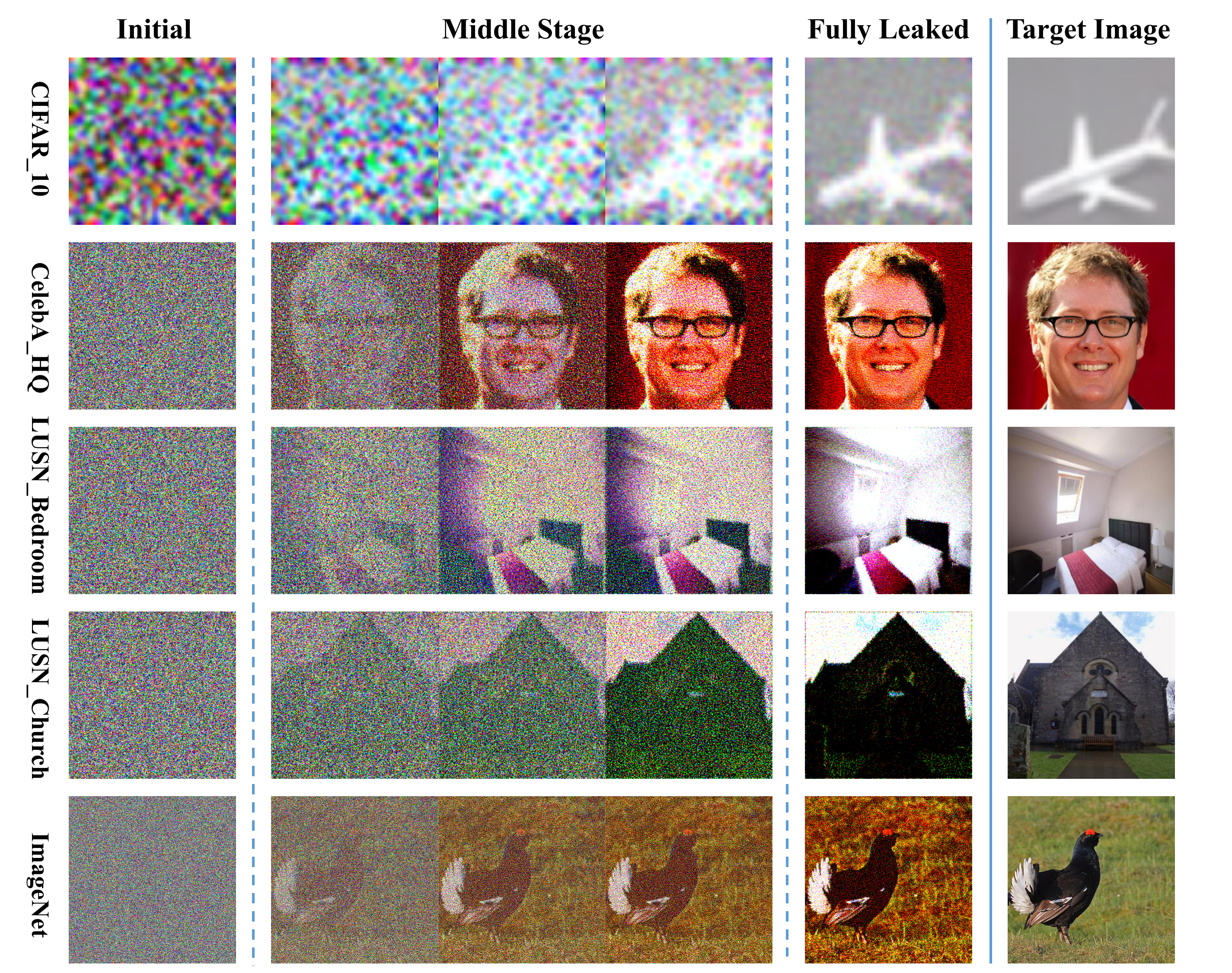}
		\label{dlg_result}
    }
    \subfigure[Gradient-Guided Diffusion Model (Ours)]{
		\includegraphics[width=0.48\textwidth]{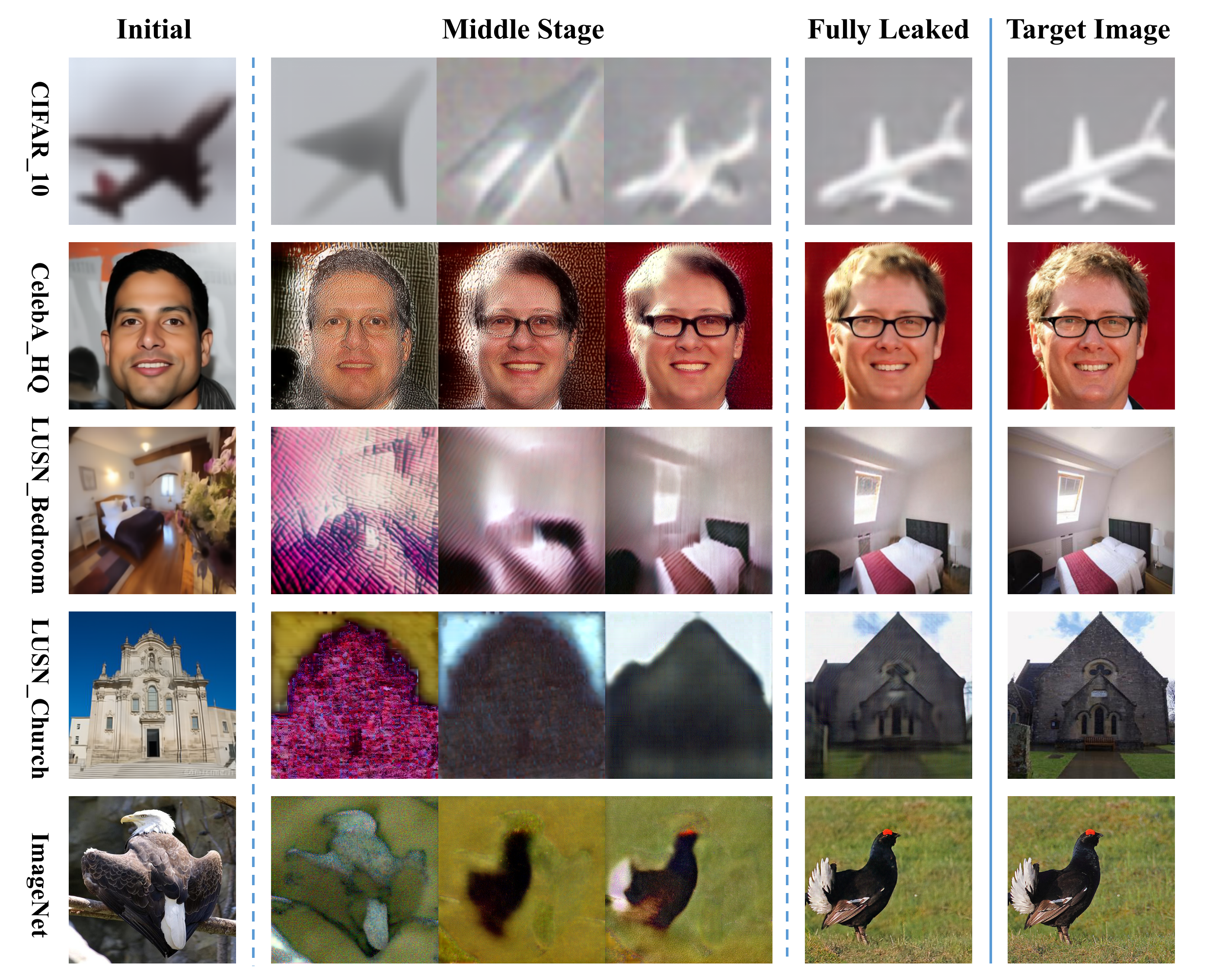}
		\label{our_result}
    }
    \caption{Reconstruction processes and results of DLG and our method.}
    \label{our_dlg}
\end{figure*}

(1) \textit{Reconstruction Studies.} The comparative results of the reconstruction processes utilizing our gradient-guided diffusion model and DLG are illustrated in Figure \ref{our_dlg}. Our method obtains fully leaked images at 200 iterations, compared to 1000 with DLG. Despite fewer iterations, our model significantly outperforms DLG in the similarity between the reconstructed images and the target ones. The quality of images reconstructed by our method remains superior to those of DLG across all datasets. 

\begin{figure*}[!htp]
  \centering
  \includegraphics[scale=0.15]{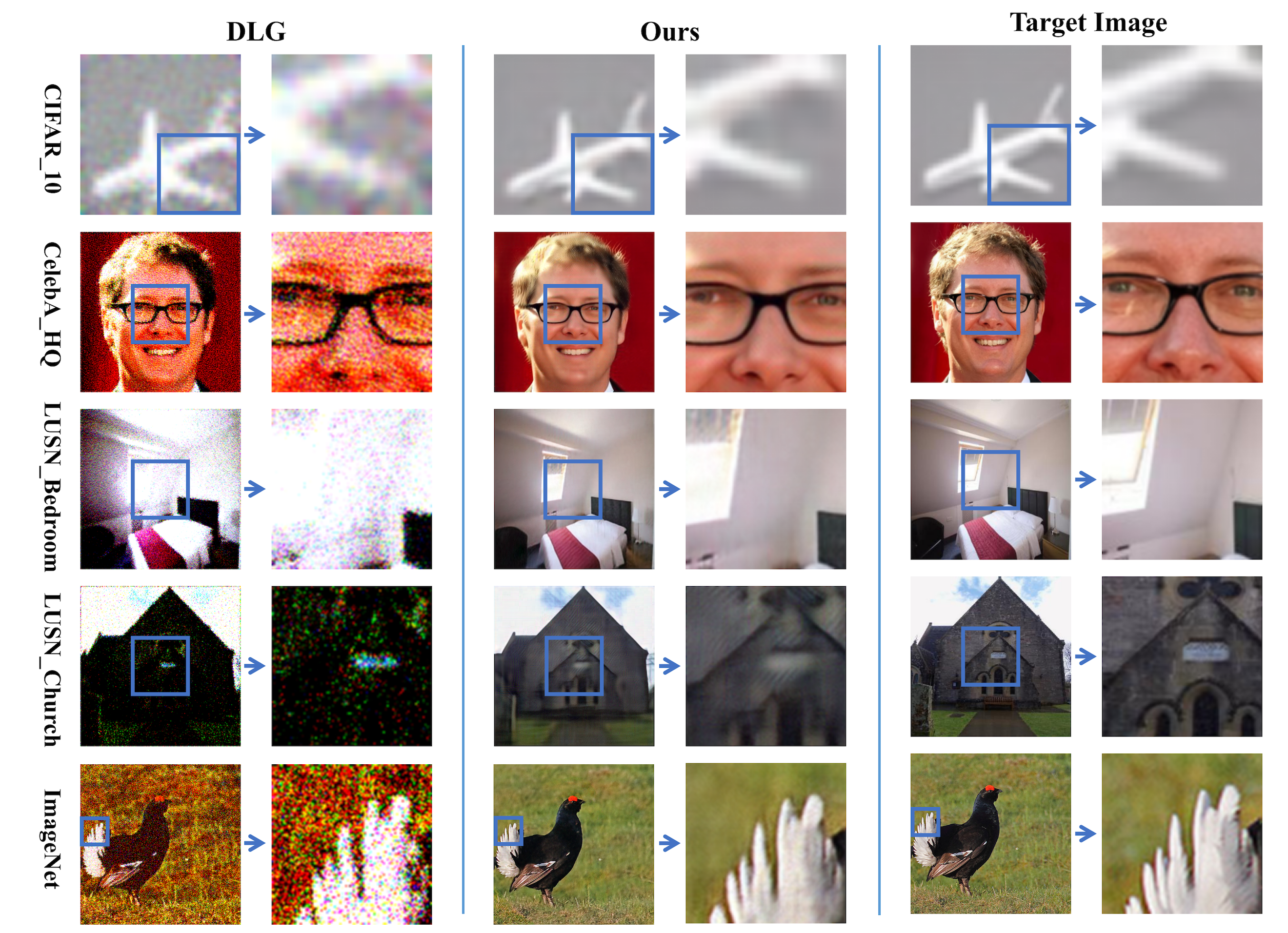}
  \caption{Comparison of the detail images of the reconstruction results via DLG and our gradient-guided diffusion model.}
  \label{xijie}
\end{figure*}

For a more detailed observation, we have zoomed in on a specific section of the fully leaked reconstruction results to compare the performance between our method and DLG in Figure \ref{xijie}. We also provide detail images of the target images, which serve as the ground truth. Despite our iterations being significantly fewer than those in DLG, the detail images of our results more closely resemble the ground truth, whereas DLG's results appear distorted. This closer inspection reveals that our method produces smoother textures with better restoration, while DLG's detail image contains many noise artifacts, leading to lower quality.

\begin{figure*}[htp!]
    \centering
    \vspace{-2mm}
    \subfigure[Deep Leakage from Gradients (DLG)]{
		\includegraphics[width=0.48\textwidth]{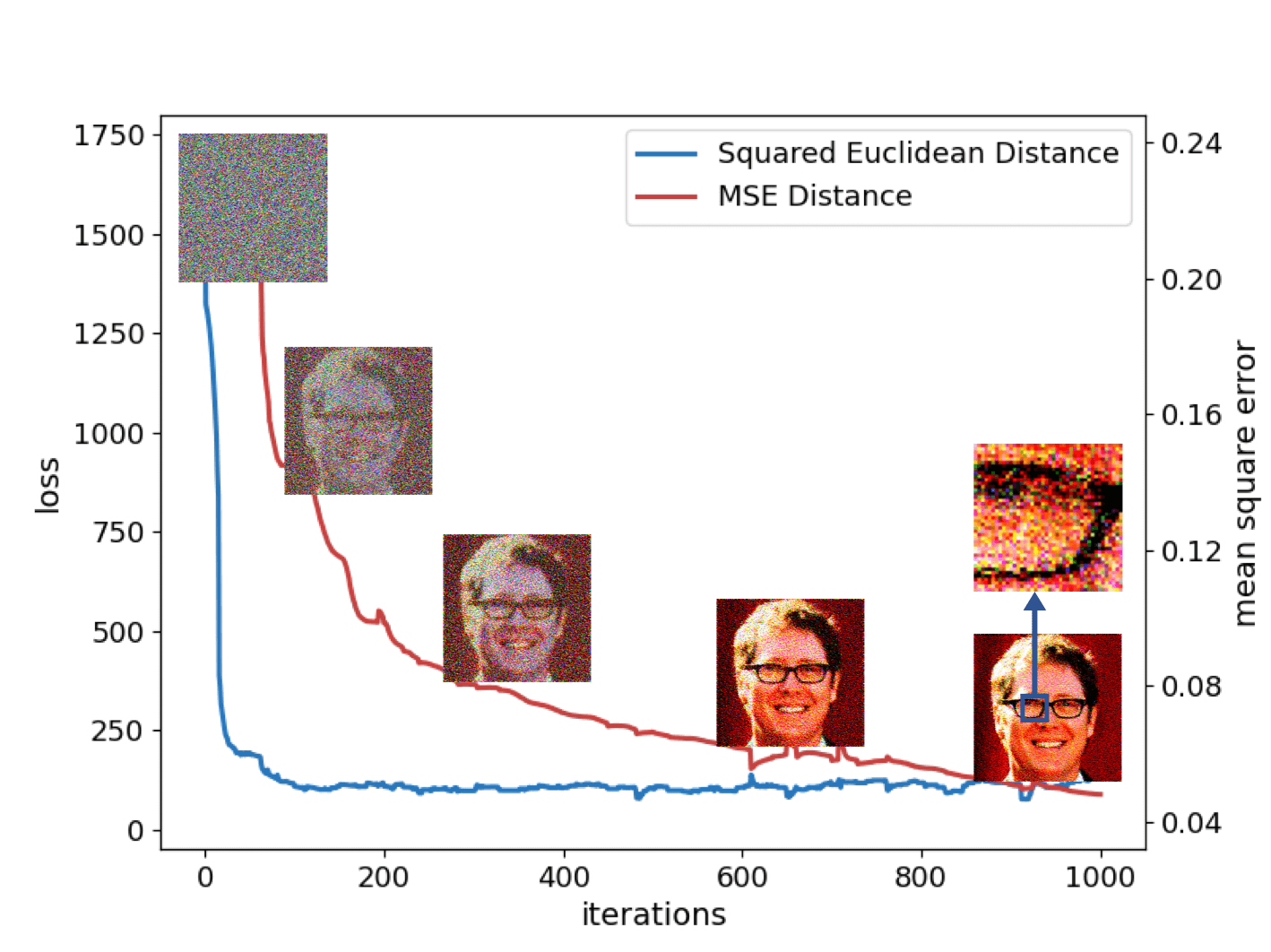}
		\label{fig4b_main}
    }
    \subfigure[Gradient-Guided Diffusion Model (Ours)]{
		\includegraphics[width=0.48\textwidth]{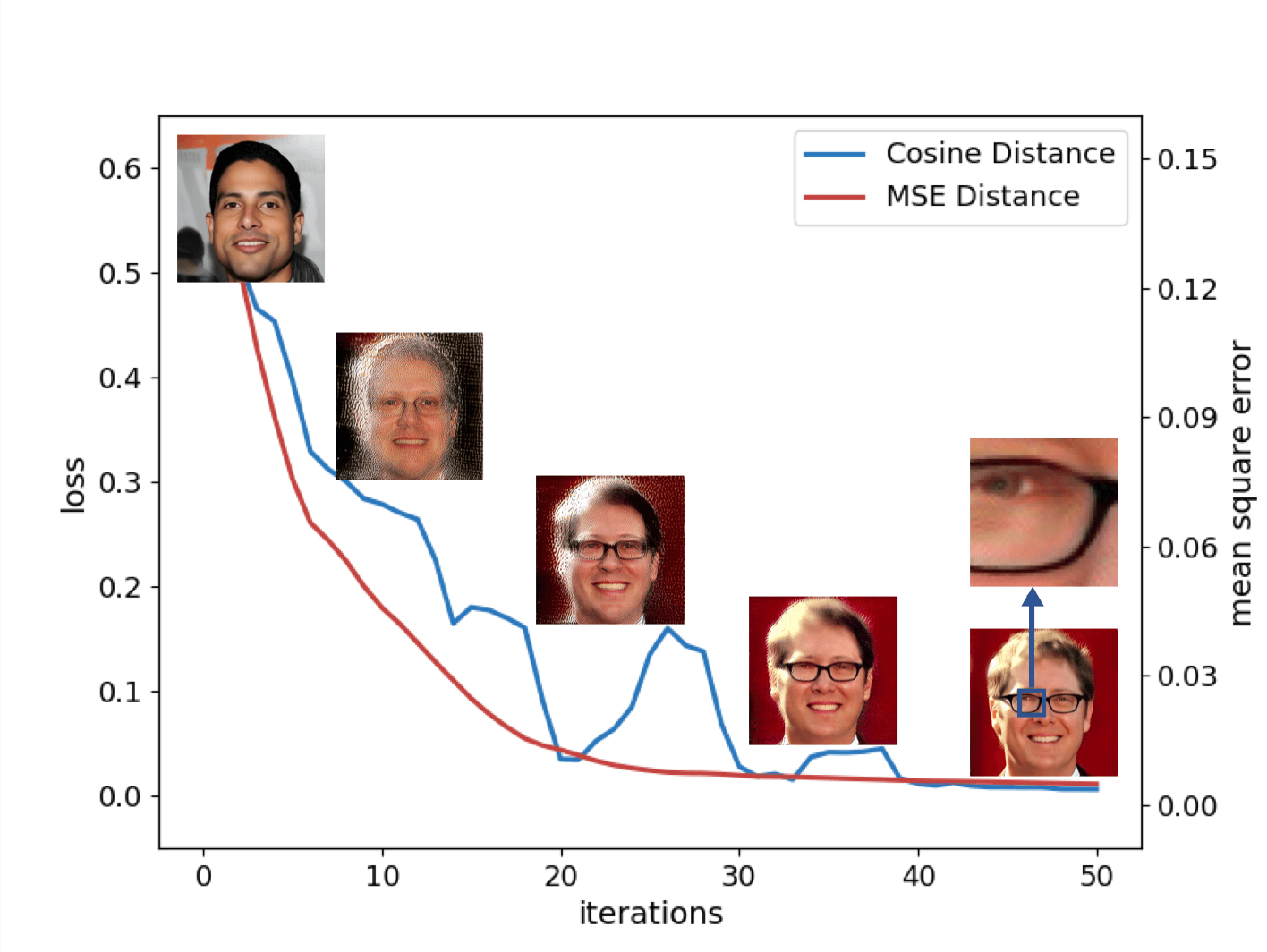}
		\label{fig4a_main}
    }
    \caption{Changing trends of reconstruction loss, Mean Squared Error (MSE) between reconstructed image and target image, as well as the appearance of images with increasing iteration on CelebA-HQ.}
    \label{fig4_main}
\end{figure*}

To facilitate understanding, we graph the changing trends of reconstruction loss, Mean Squared Error (MSE) between the reconstructed image and target image, and the appearance of images with increasing iteration on CelebA-HQ in Figure \ref{fig4_main}. Specifically, the reconstruction losses for our method and DLG are quantified using cosine distance and squared Euclidean distance of gradients, respectively. The figures indicate that for both methods, a decline in reconstruction loss is paralleled by a decrease in MSE distance. Remarkably, in our method, the MSE values effectively converge at around 30 iterations, and the reconstructed image then becomes visually indistinguishable from the target image to the human eye. This represents a substantial enhancement over DLG, which necessitates approximately 1000 iterations to achieve convergence. Furthermore, we showcase an enlarged view of specific areas from the reconstruction results at the highest iteration in figures (50 in our method, and 1000 in DLG), which further validates the effectiveness of our attack in terms of image smoothness and restoration.

(2) \textit{Quantitative Evaluations.} The results of our quantitative evaluations are presented in Table~\ref{table1}. We configure our gradient-guided fine-tuning iteration to 200, in contrast to the 1000 iterations required by DLG. For a more challenging attempt, because of the high resolution of images in ImageNet, we wonder whether we can reconstruct them by fewer reverse steps $S_{gen}$ to save GPU consumption. Thus, we set $S_{gen}$ to 2 for ImageNet. Our method demonstrates significantly improved performance, as evidenced by substantially lower MSE and LPIPS scores, alongside notably higher SSIM and PSNR values. This demonstrates that the image reconstruction quality using our gradient-guided diffusion model markedly surpasses that achieved with DLG. Additionally, our method enhances efficiency in image reconstruction, necessitating fewer fine-tuning iterations and at most 6 steps in the reverse process of the fine-tuned diffusion model.
\begin{table*}[htp!]
   \caption{Quantitative evaluations of DLG and our method.}
   \label{table1}
   \centering
   \scalebox{1}{
        \begin{tabular}{c|cc|cc|cc|cc}
            \hline
            \toprule
            \textbf{\multirow{2}{*}{Dataset}} & \multicolumn{2}{c|}{\textbf{MSE$\downarrow$}} & \multicolumn{2}{c|}{\textbf{SSIM$\uparrow$}} & \multicolumn{2}{c|}{\textbf{PSNR$\uparrow$}} & \multicolumn{2}{c}{\textbf{LPIPS$\downarrow$}}                                       \\ \cline{2-9}
             & DLG   & \textbf{Ours}  & DLG   & \textbf{Ours}   & DLG   & \textbf{Ours}   & \multicolumn{1}{c}{DLG} & \textbf{Ours}                      \\ \midrule
            \textbf{CIFAR\_10}                                         & 0.0054      & \textbf{0.0004}     & 0.9998      & \textbf{1.0000}      & 22.6747     & \textbf{33.6235}     & \multicolumn{1}{c}{3.3553e-07}    & \textbf{3.7686e-08}                         \\ 
            \textbf{CelebA-HQ}                                        & 0.0480      & \textbf{0.0030}     & 0.9979      & \textbf{0.9999}      & 13.1835      & \textbf{25.2740}     & \multicolumn{1}{c}{2.3453e-04}    & \textbf{2.9193e-05}                         \\ 
            \textbf{LSUN\_Bedroom}                                     & 0.0522      & \textbf{0.0006}     & 0.9953      & \textbf{1.0000}      & 12.8265      & \textbf{32.4784}     & \multicolumn{1}{c}{2.3804e-04}    & \textbf{3.6974e-06}                         \\
            \textbf{LSUN\_Church}                                      & 0.0481      & \textbf{0.0025}     & 0.9943      & \textbf{0.9999}     & 13.1816      & \textbf{26.0303}     & \multicolumn{1}{c}{1.0334e-04}    & \textbf{1.7650e-05}                         \\ 
            \textbf{ImageNet}                                      & 0.0707      & \textbf{0.0030}     & 0.9977      & \textbf{0.9999}     & 11.4573      & \textbf{25.2863}     & \multicolumn{1}{c}{5.9084e-04}    & \textbf{1.3781e-05} \\ \bottomrule
        \end{tabular}
    }
\end{table*}
\subsubsection{Comparison with Stealing Capability of Additional Baselines}

To thoroughly show our method's superior performance, we compare our method with the additional state-of-the-art (SOTA) baselines mentioned in Subsection \ref{subsection2.1}: \textit{(1) Inverting Gradients (IG) by \cite{ig}; (2) Grad-Inversion (GI) by \cite{gi}; (3) Gradient Inversion in Alternative Spaces (GIAS) by \cite{gias}; (4) Generative Gradient Leakage (GGL) by \cite{ggl}; (5) Generative Gradient Inversion Method with Feature Domain Optimization (GIFD) by \cite{gifd}.} All these baselines have conducted experiments on ImageNet, so our method can be directly compared with them. We replicate the baseline experiments and adhere to the settings as described in \cite{gifd}, where evaluating performances on $64 \times 64$ pixels images. When conducting our method, due to limitations imposed by the resolution of images in the training dataset used for our pre-trained diffusion model, we maintain the same settings as detailed in Subsection \ref{subsection.4.1} and reconstruct images with the resolution of $512 \times 512$ pixels, which is a more challenging task. 

\begin{figure*}[!htp]
  \centering
  \includegraphics[scale=0.13]{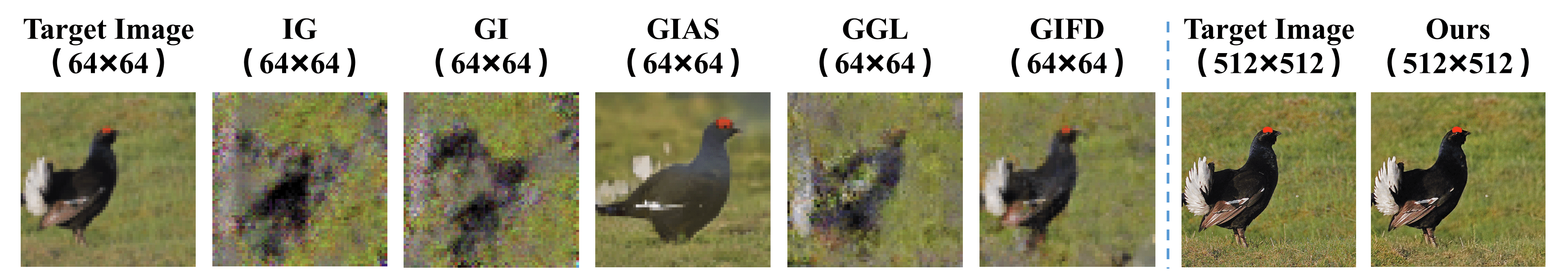}
  \caption{Reconstruction results of additional SOTA baselines and our method.}
  \label{baselines}
\end{figure*}

(1) \textit{Reconstruction Studies.} As illustrated in Figure \ref{baselines}, experiments are conducted on the same target image at different resolutions, where these advanced baselines use the $64 \times 64$ pixels image while our method uses the $512 \times 512$ pixels image. It is evident that, despite our method being applied to a more challenging task, the visual reconstruction quality is significantly better than the baselines.

\begin{table*}[!htp]
   \caption{Quantitative evaluations on ImageNet of additional SOTA baselines and ours.}
   \label{table_high}
   \centering
   \scalebox{1}{
        \begin{tabular}{c|cccccc}
            \hline
            \toprule
            \textbf{\multirow{2}{*}{Metrics}} & \textbf{IG} & \textbf{GI} & \textbf{GIAS} & \textbf{GGL} & \textbf{GIFD} & \textbf{Ours}   \\ 
 & \textbf{$(64 \times 64)$} & \textbf{$(64 \times 64)$} & \textbf{$(64 \times 64)$} & \textbf{$(64 \times 64)$} & \textbf{$(64 \times 64)$} & \textbf{$(512 \times 512)$}   \\ \midrule
            \textbf{MSE$\downarrow$} & 0.0196 & 0.0223 & 0.0458 & 0.0179 & 0.0098 & \textbf{0.0030} \\ 
            \textbf{SSIM$\uparrow$} & 0.9982 & 0.9978 & 0.9953 & 0.9987 & 0.9991 & \textbf{0.9999} \\ 
            \textbf{PSNR$\uparrow$} & 17.0756 & 16.5109 & 13.3885 & 17.4923 & 20.0534 & \textbf{25.2863} \\
            \textbf{LPIPS$\downarrow$} & 1.1612e-04 & 1.4274e-04 & 3.9351e-04 & 1.1937e-04 & 6.6672e-05 & \textbf{1.3781e-05} \\ \bottomrule
        \end{tabular}
    }
\end{table*}

(2) \textit{Quantitative Evaluations.} The quantitative results are presented in Table \ref{table_high}. The baselines reconstruct the $64 \times 64$ pixels image, whereas our method reconstructs the image up to a resolution of $512 \times 512$ pixels. Table \ref{table_high} reveals that our method's metrics outperform all other baselines. This highlights the superior performance of our method in terms of image reconstruction effectiveness.

\subsubsection{Comparison of Time Efficiency}

Regarding time efficiency, Figure \ref{fig4a_main} demonstrates that our method converges in roughly 30 iterations. According to Appendix \ref{appendix.d.1.}, our method takes about 130 seconds to converge when reconstructing images of $256 \times 256$ pixels. Compared to the most time efficient baseline mentioned in \cite{gifd}, IG, which needs around 181 seconds to converge on images of only $64 \times 64$ pixels, our method is significantly more time efficient. For more details please refer to Appendix \ref{appendix.d}.

\subsection{Noisy Gradients}

One straightforward attempt to defend against image reconstruction attacks is to add noise on gradients before sharing. To evaluate, we conduct experiments using Gaussian and Laplacian noises (widely used in differential privacy \cite{dp} studies) with variance range from $10^{-1}$ to $10^{-4}$ and mean 0.

\begin{figure*}[htp!]
    \centering
    \vspace{-2mm}
    \subfigure[Deap Leakage from Gradients (DLG)]{
		\includegraphics[width=0.45\textwidth]{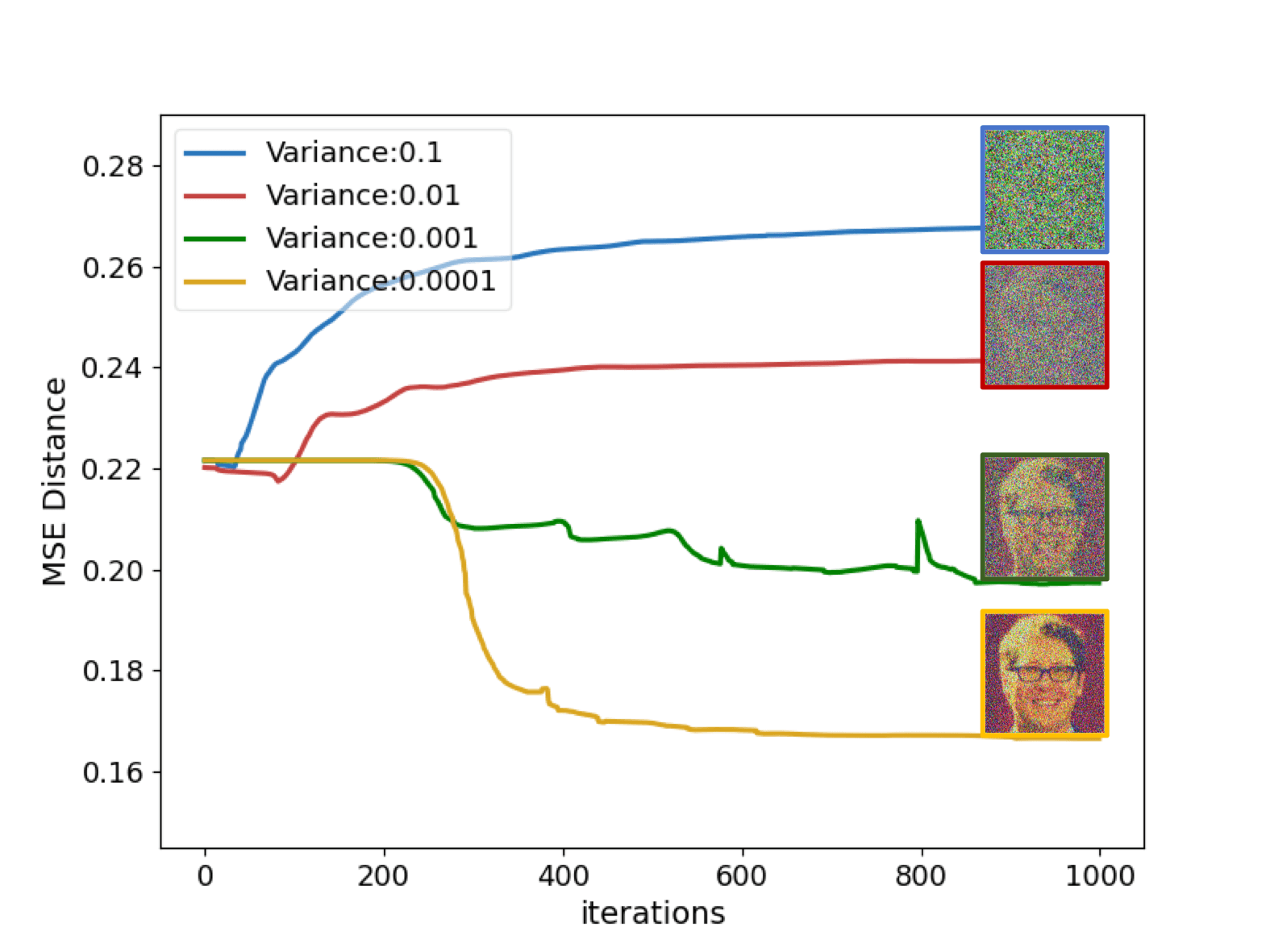}
		\label{noise_gau_dlg}
    }
    \subfigure[Gradient-Guided Diffusion Model (Ours)]{
		\includegraphics[width=0.45\textwidth]{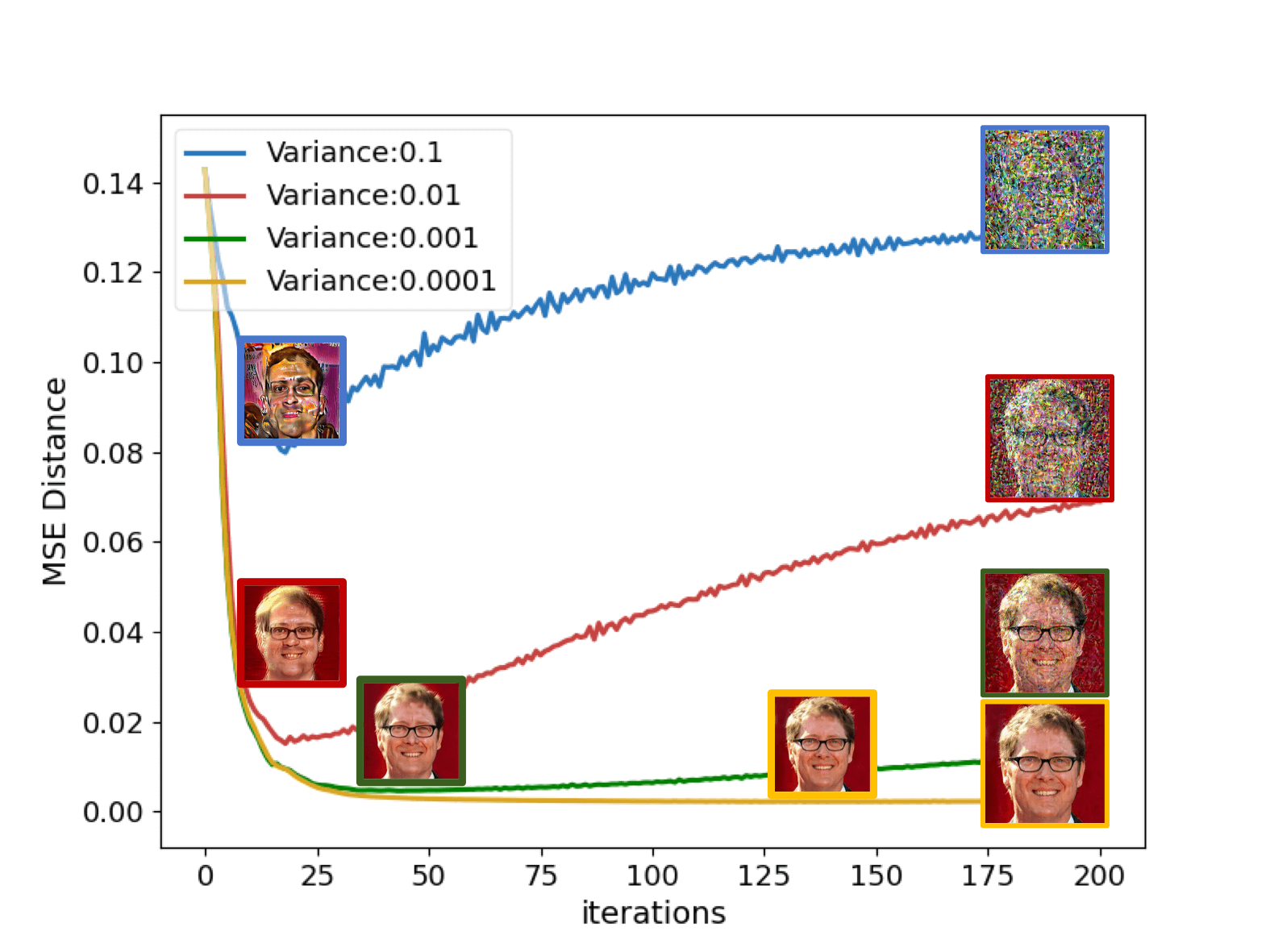}
		\label{noise_gau_ours}
    }
    \caption{Changing trends of Mean Squared Error (MSE) between reconstructed image and target image, as well as the appearance of images under defenses with Gaussian noise on CelebA-HQ. Specifically, we display images at the lowest MSE in our method, marking peak performance.}
    \label{noise}
\end{figure*}

\begin{figure*}[htp!]
    \centering
    \vspace{-2mm}
    \subfigure[Deap Leakage from Gradients (DLG)]{
		\includegraphics[width=0.45\textwidth]{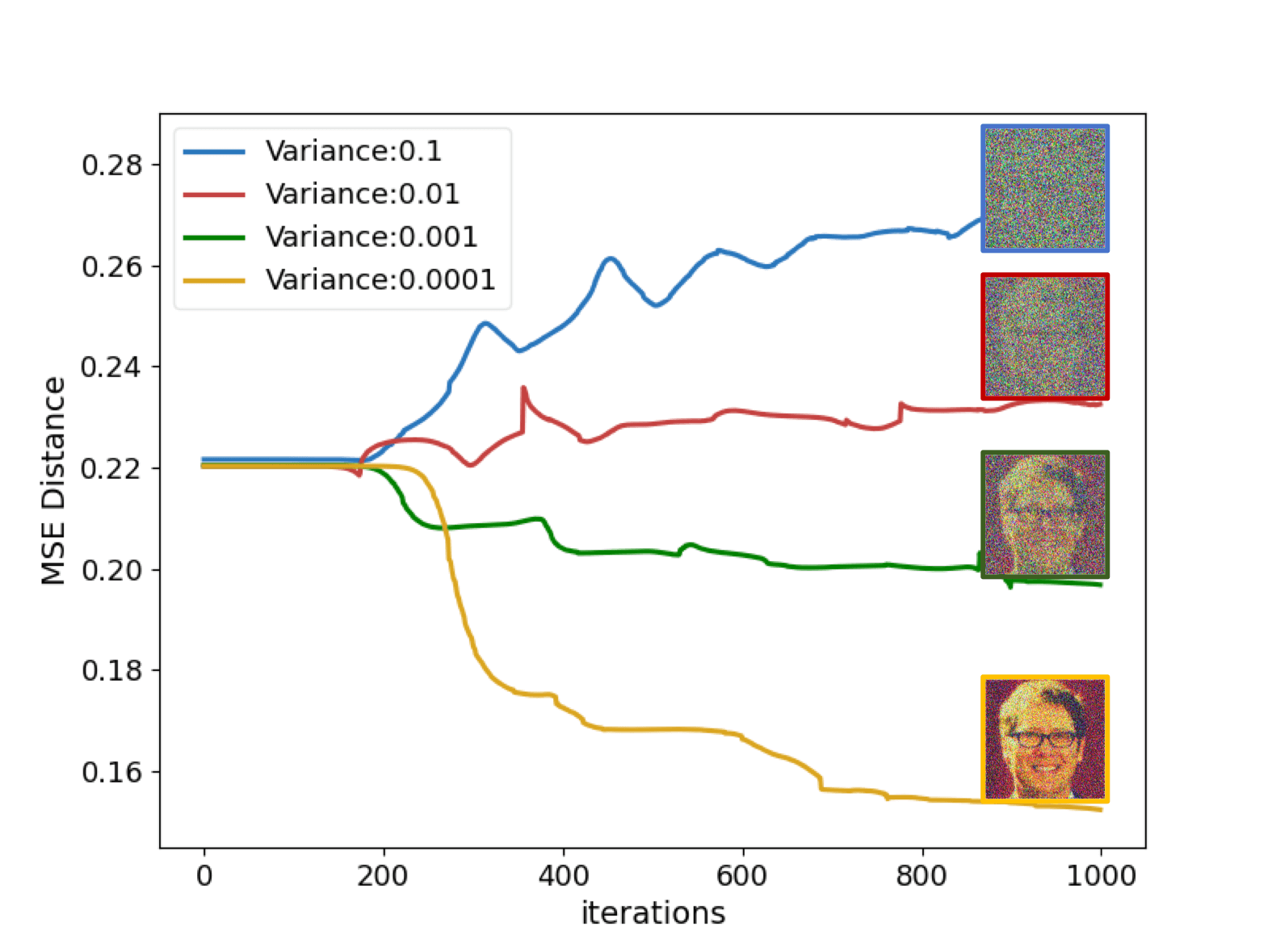}
		\label{noise_lap_dlg_appendix}
    }
    \subfigure[Gradient-Guided Diffusion Model (Ours)]{
		\includegraphics[width=0.45\textwidth]{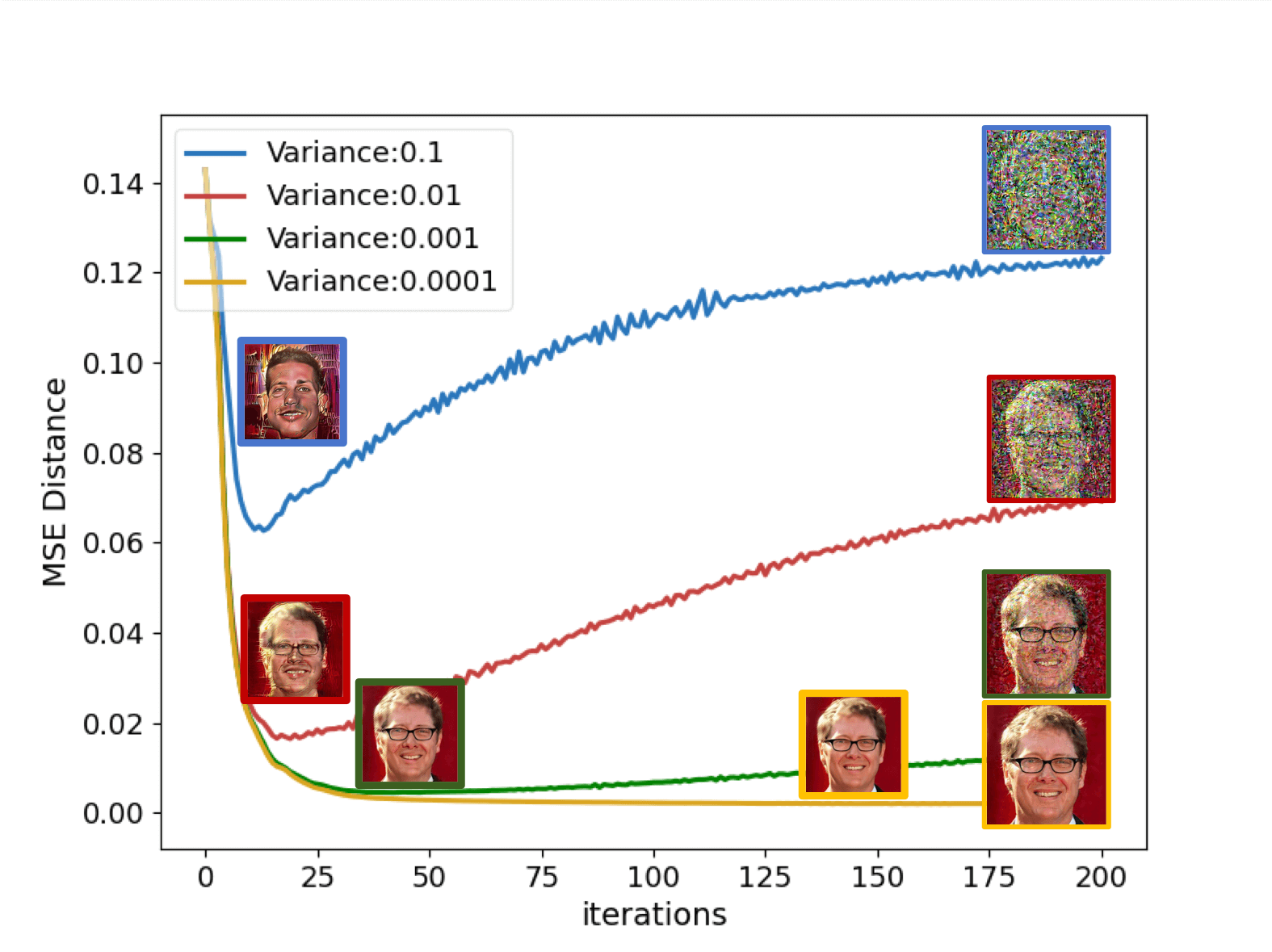}
		\label{noise_lap_ours_appendix}
    }
    \caption{Changing trends of Mean Squared Error (MSE) between reconstructed image and target image, as well as the appearance of images under defenses with Laplacian noise on CelebA-HQ. Specifically, we display images at the lowest MSE in our method, marking peak performance.}
    \label{noise_lap_appendix}
\end{figure*}

(1) \textit{Reconstruction Studies.} Figure \ref{noise} and Figure \ref{noise_lap_appendix} show MSE between the reconstructed image via DLG and the target image under different magnitude noise. We can find that the quality of the reconstruction image primarily depends on the magnitude of distribution variance and the attack method, and is less influenced by the type of noise. The MSE of DLG gradually decreases and converges with low noise levels. Unsurprisingly, when the variance is larger than $10^{-2}$, MSE increases instead of decreasing, and the reconstruction results of DLG are even worse than completely noisy images. Differential privacy successfully blocked attack of DLG. Compared to DLG, as iteration increases, the MSE of our method initially decreases before rising. We denote the reconstructed image concerning the lowest MSE as the peak performance. The results indicate that with noisy gradients, our method has the optimal results at the peak performance. At the maximum iteration, images exhibit more noise artifacts due to noise incorporated in the gradients, which leads the subsequent increase of MSE. When variance is at the scale of $10^{-4}$ and $10^{-3}$, the noisy gradients can not prevent our attack. For noise with variance $10^{-2}$, although the peak performance shows some differences with the target image, we can still extract important private information. For example, we can determine that the target image shows a man wearing glasses, smiling with teeth visible. The leakage can still be performed. Only when the variance is larger than $10^{-1}$, the noise is starting to affect the accuracy.

\begin{figure*}[htp!]
    \centering
    \vspace{-2mm}
    \subfigure[Deap Leakage from Gradients (DLG)]{
		\includegraphics[height=0.22\textheight]{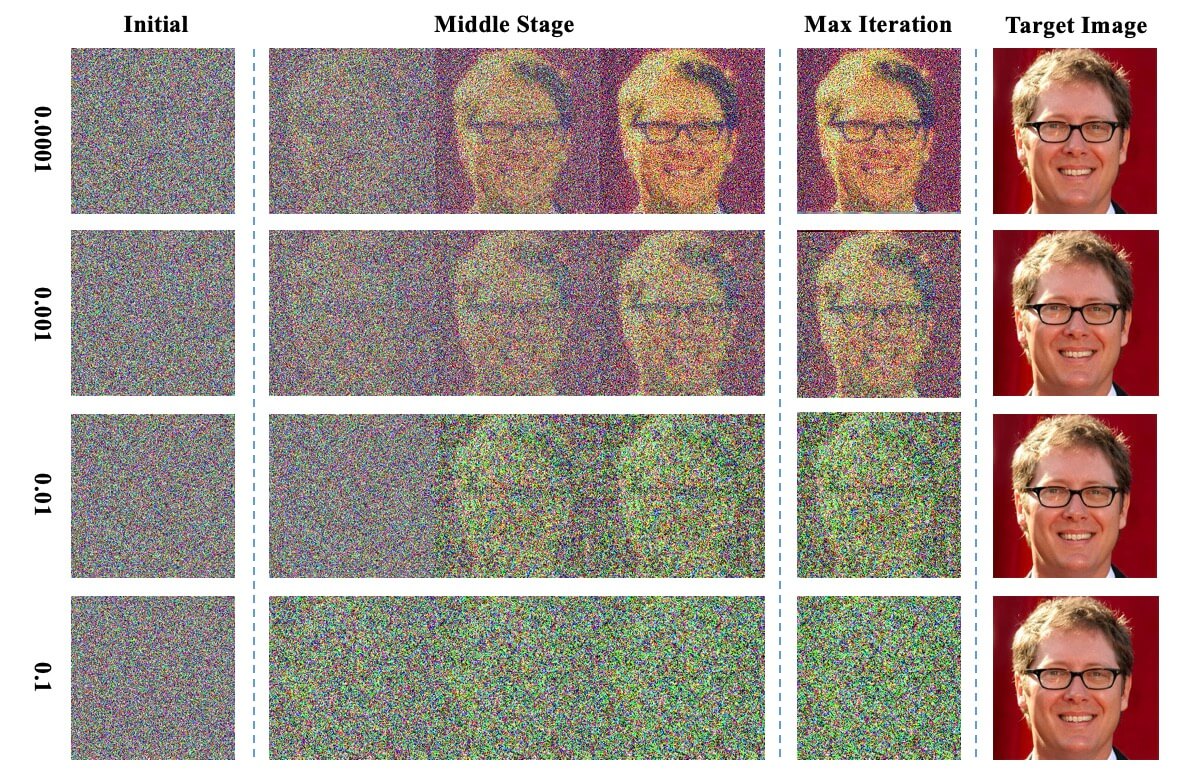}
		\label{noise_gauu}
    }
    \subfigure[Gradient-Guided Diffusion Model (Ours)]{
		\includegraphics[height=0.22\textheight]{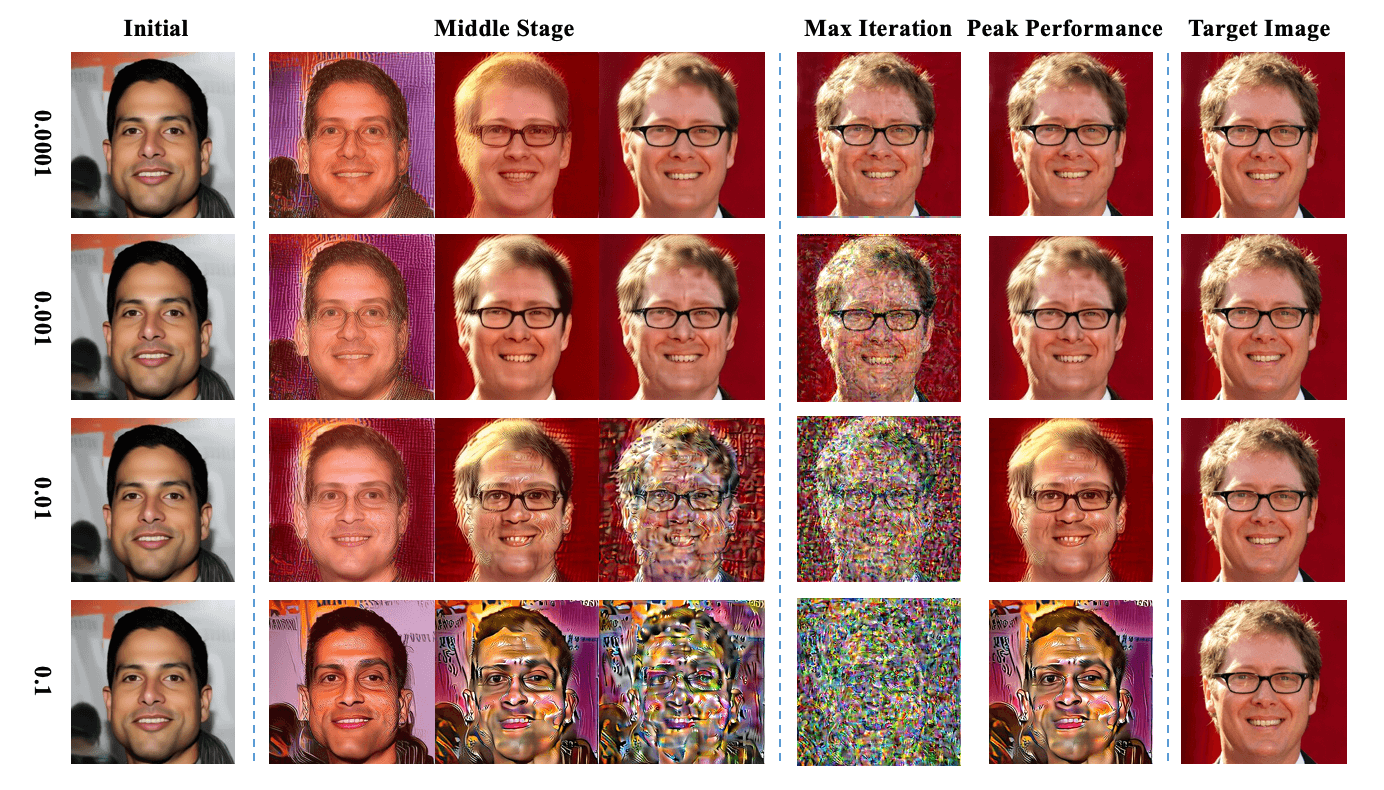}
		\label{noise_lappp}
    }
    \caption{Reconstruction processes under defenses with Gaussian noise on CelebA-HQ. The maximum iteration of DLG is 1000, and the maximum iteration of our method is 200.}
    \label{noise_gau_process}
\end{figure*}

\begin{figure*}[htp!]
    \centering
    \vspace{-2mm}
    \subfigure[Deap Leakage from Gradients (DLG)]{
		\includegraphics[height=0.22\textheight]{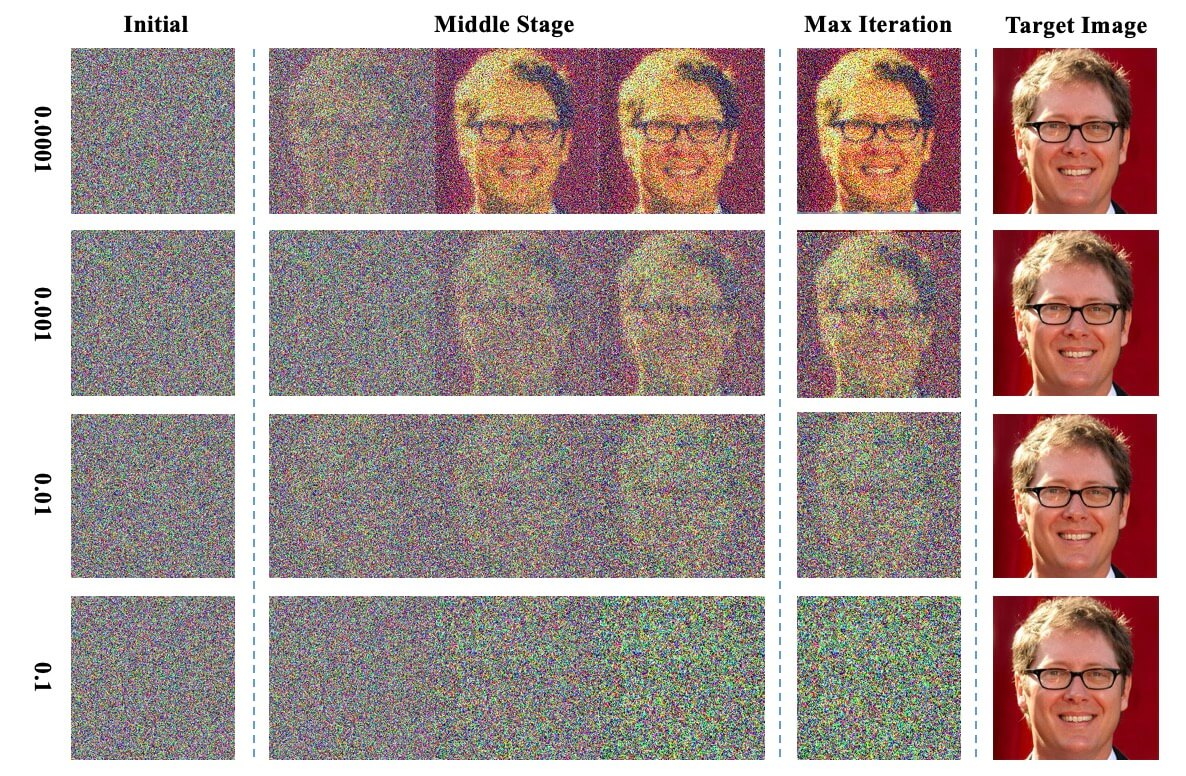}
		\label{noise_gauuu}
    }
    \subfigure[Gradient-Guided Diffusion Model (Ours)]{
		\includegraphics[height=0.22\textheight]{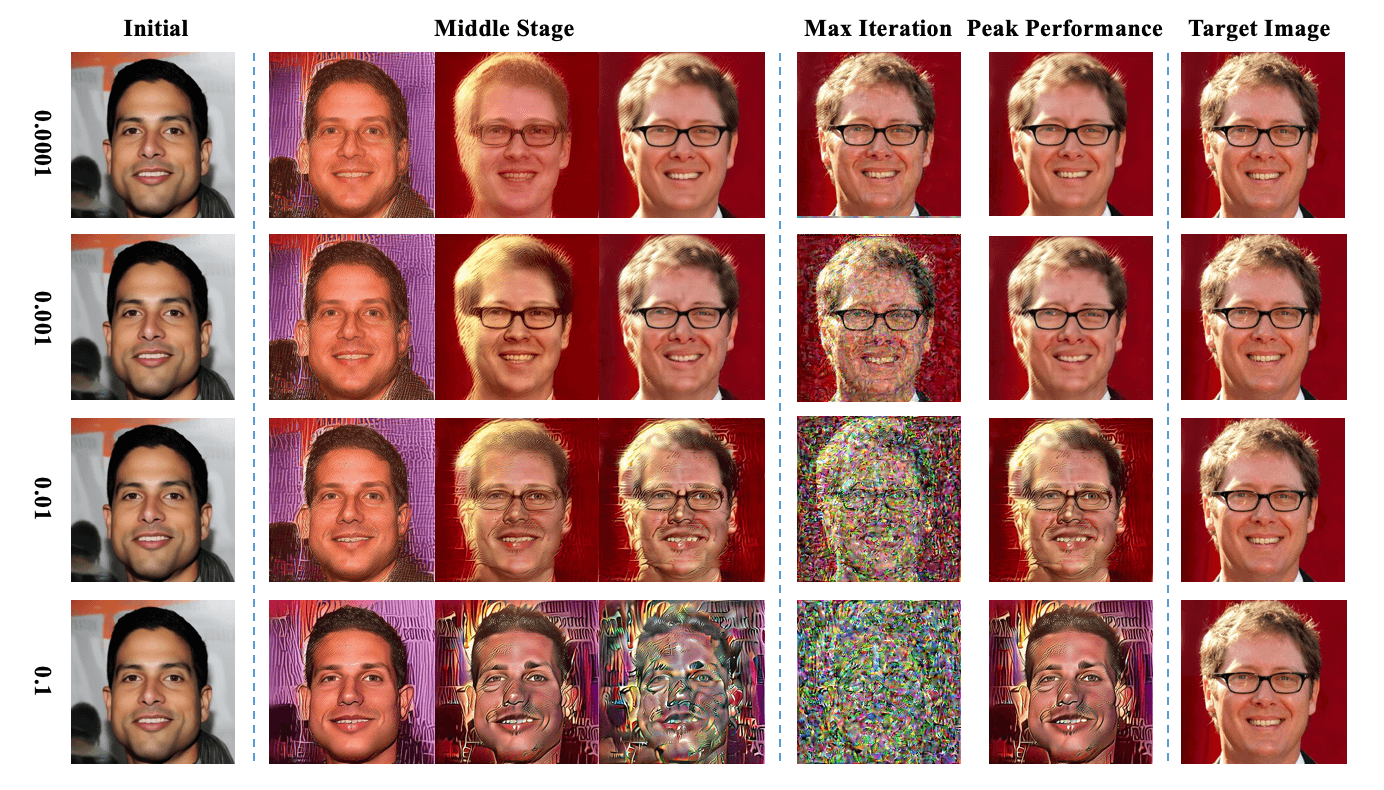}
		\label{noise_lapp}
    }
    \caption{Reconstruction processes under defenses with Laplacian noise on CelebA-HQ. The maximum iteration of DLG is 1000, and the maximum iteration of our method is 200.}
    \label{noise_lap_process}
\end{figure*}

In addition, we present images during the reconstruction process of DLG and our method. As Figure ~\ref{noise_gau_process} and Figure \ref{noise_lap_process} show, during the fine-tuning process, the images steadily converge in DLG. With diffusion model, our method initially reconstructed the smooth private image more accurately, followed by the appearance of more noisy pixels. This more intuitively explains the trend of MSE first decreasing and then increasing in our method.

(2) \textit{Quantitative Evaluations.} The quantitative results are presented in Table \ref{defense}, which further confirms that our method's attack is superior, even with noisy target gradients.

\begin{table*}[!htb]
    \caption{Quantitative analysis on reconstruction quality with noisy gradients of DLG and our method. The similarity metrics are between the image at peak performance and the target image.}
    \label{defense}
    \centering
    \begin{tabular}{c|c|ccccc}
    \hline
    \toprule
      & \textbf{Noise Type} & \textbf{Variance} & \textbf{MSE$\downarrow$} & \textbf{SSIM$\uparrow$} & \textbf{PSNR$\uparrow$}    & \textbf{LPIPS$\downarrow$} \\
    \midrule
    \multirow{8}{*}{DLG} &\multirow{4}{*}{Gaussian} & $10^{-4}$ &0.1619  &0.9942 & 7.9068& 1.9422e-03\\
                         &  & $10^{-3}$ &0.1971 &0.9908&7.0528 & 1.4500e-03 \\
                         &  & $10^{-2}$ & 0.2194&0.9854 &6.5871& 1.8540e-03\\
                         &  & $10^{-1}$ & 0.2203& 0.9851&6.5695 & 1.4226e-03\\ \cline{2-7}
                         &\multirow{4}{*}{Laplacian} & $10^{-4}$ & 0.1510 & 0.9952 & 8.2112& 1.0127e-03\\
                         &  & $10^{-3}$ & 0.1963&0.9907&7.0716 & 1.7223e-03 \\
                         &  & $10^{-2}$ &0.2184 &0.9855&6.6069 & 1.4013e-03\\
                         &  & $10^{-1}$ &0.2214 &0.9851& 6.5478& 1.3835e-03\\ \midrule
    \multirow{8}{*}{Ours} &\multirow{4}{*}{Gaussian} & $10^{-4}$ & 0.0086 & 0.9997 & 20.6627 & 8.2618e-06\\
                         &  & $10^{-3}$& 0.0185 & 0.9990  & 17.3366&2.7708e-05\\
                         &  & $10^{-2}$ &  0.0681 & 0.9947 & 11.6675 & 1.0877e-04\\
                         &  & $10^{-1}$ & 0.0798 & 0.9924 &10.9788 &5.6060e-04\\ \cline{2-7}
                         &\multirow{4}{*}{Laplacian} &$10^{-4}$ & 0.0079 & 0.9997 & 21.0407 & 7.0692e-06\\
                         &  & $10^{-3}$&0.0178 &0.9991&17.4850 &2.1464e-05\\
                         &  & $10^{-2}$ & 0.0688&0.9951& 11.6253 &1.5671e-04 \\
                         &  & $10^{-1}$ &0.1150 & 0.9887 & 9.3913 &3.9854e-04 \\
    \bottomrule
    \end{tabular}
\end{table*}

In conclusion, differential privacy can completely defend against DLG, but additional noise is required to counteract attacks using diffusion models. Our significant discovery is that reconstruction attacks based on diffusion models have the optimal reconstruction of the target images during the intermediate process. This may be because the impact of the noise added to the gradients on the reconstructed images is mitigated to some extent when the dimension of the latent space is lowered in the U-net of the reverse process of the diffusion model, thereby enhancing attack's ability. However, DLG encounters failures during the whole reconstruction period. Our method exhibits some resilience to differential privacy, which will be investigated in future work.

\subsection{Hyperparameter and Ablation Study} 

\begin{figure*}[!htp]
  \centering
  \includegraphics[scale=0.3]{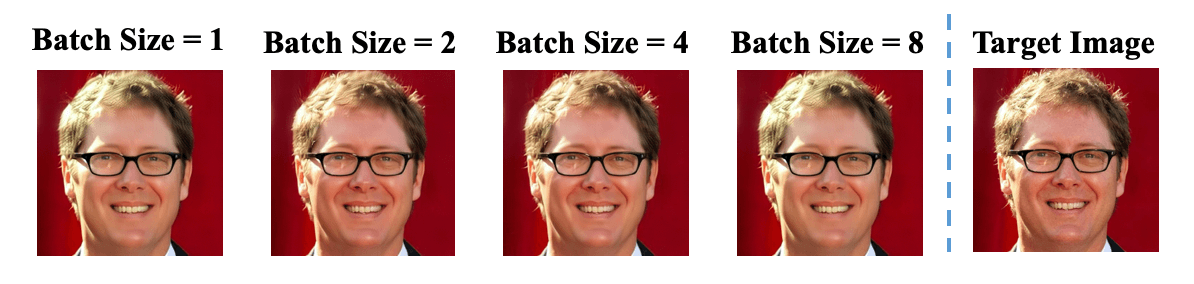}
  \caption{Dependency of reconstruction results on batch size. All images are fully leaked where maximum number of iterations equals 200.}
  \label{batch}
\end{figure*}

\begin{table*}[!htb]
    \caption{Quantitative analysis on reconstruction quality with respect to Adam's parameters: initial learning rate and decay rate.}
    \label{table2}
    \centering
    \begin{tabular}{c|c|ccrc}
    \hline
    \toprule
    \textbf{Initial Learning Rate}       & \textbf{Decay rate}    & \multicolumn{4}{c}{\textbf{Evaluation Metrics}} \\ \cline{3-6} 
    \textbf{$\eta$}       &   \textbf{$\gamma$}   & \textbf{MSE$\downarrow$} & \textbf{SSIM$\uparrow$} & \textbf{PSNR$\uparrow$}    & \textbf{LPIPS$\downarrow$} \\
    \midrule
    \multirow{6}{*}{2e-05} & 0.90 & 0.0177 & 0.99877 & 17.5238 & 1.1825e-04 \\
     & 0.94 & 0.0059 & 0.99970 & 22.2980 & 3.4723e-05 \\
     & 0.98 & 0.0028 & 0.99989 & 25.5877 & 2.2213e-05 \\
     & 1.02 & 0.0020 & 0.99992 & 26.9616 & 1.1547e-05 \\
     & 1.06 & 0.2317 & 0.96620 & 6.3506 & 4.3700e-04 \\
     & 1.10 & 0.2317 & 0.96620 & 6.3506 & 4.3700e-04 \\ \cline{1-6}
    1e-04 & \multirow{6}{*}{0.98} & 0.0110 & 0.99922 & 19.5976 & 1.0627e-04 \\
    6e-05 &  & 0.0042 & 0.99976 & 23.7413 & 6.7694e-05 \\
    2e-05 &  & 0.0028 & 0.99989 & 25.5877 & 2.2213e-05 \\
    8e-06 &  & 0.0110 & 0.99919 & 19.5721 & 8.0014e-05 \\
    4e-06 &  & 0.0074 & 0.99941 & 21.3243 & 3.7889e-05 \\
    8e-07 &  & 0.0807 & 0.99239 & 10.9334 & 5.5601e-04 \\
    \bottomrule
    \end{tabular}
\end{table*}

We present partial results of ablation study, please refer to the Appendix \ref{appendix_ablation} for more results.

\subsubsection{Settings of Adam Optimizer}
Here, we investigate the impact of the initial learning rate and the decay rate of learning rate of the Adam optimizer on the reconstruction performance of our method on CelebA-HQ. As presented in Table~\ref{table2}, we first fix the initial learning rate and vary the decay rate. The results indicate that as the decay rate increases, the reconstruction performance initially improves and then deteriorates. When the decay rate is less than 1 and too small, the learning rate decreases rapidly with increasing iteration. Consequently, after a few iterations, the learning rate diminishes to a small value, significantly slowing down the optimization process. This hinders optimizer's ability to rapidly optimize the reconstruction performance. Conversely, if the decay rate exceeds 1, the learning rate will gradually increase as the number of iterations grows. However, if the decay rate is excessively large, the learning rate will increase to an overly high value, leading to model instability during the training process. An excessively high learning rate can cause significant fluctuations in the loss function, thereby impeding convergence to the optimal solution. Therefore, selecting an appropriately moderate decay rate, typically around 1, is crucial. On the other hand, when we keep the value of the decay rate fixed at 0.98 as the settings of our experiments, it is essential to ensure that the initial learning rate should be set neither too large nor too small. Our experiment results indicate that the optimal initial learning rate is 2e-05. An excessively small initial learning rate significantly slows down the optimization process, necessitating more iterations to achieve comparable performance to that achieved with a moderately larger initial learning rate. Conversely, if the initial learning rate is somewhat too large, the reconstruction results after 200 iterations are suboptimal, primarily due to the instability caused by the excessively high learning rate. Therefore, maintaining a balanced initial learning rate is essential to ensure both efficiency and stability in the training process.

\subsubsection{Batch Size} 

We provide the results of using different batch sizes in fine-tuning the diffusion model to attack private images on CelebA-HQ. We keep the maximum number of iterations fixed at 200, aligning with the default setting outlined in the paper, ensuring that images are fully leaked. As shown in Table \ref{batchh}, with the increasing batch size, MSE, SSIM and PSNR slightly worsen, while LPIPS demonstrates an improvement followed by a deterioration. MSE, SSIM, and PSNR primarily focus on the local details, structural similarity, and overall fidelity of the images, and they are all based on pixel-level comparisons. LPIPS emphasizes the perceptual similarity of images, which is the overall similarity perceived by the human eye. This trend may stem from the impact of batch processing. As the batch size increases, the model reconstructs multiple images simultaneously, resulting in reduced pixel-level reconstruction quality of individual images, which is reflected in metrics such as MSE, SSIM, and PSNR. However, the LPIPS focuses more on perceptual similarity to the human eye rather than pixel-level differences. Despite some noise or distortion at pixel-level, increasing the batch size appropriately might lead to the model better capturing the perceptual similarity between images, leading to the improvement of LPIPS. Overall, the changes and fluctuations of metrics are minor. As shown in Figure \ref{batch}, there is no significant change in image quality with increasing batch size, and privacy images can still be leaked with high fidelity.

\begin{table}[H]
    \caption{Quantitative analysis on reconstruction quality with respect to different batch sizes.}
    \label{batchh}
    \centering
    \begin{tabular}{ccccc}
    \hline
    \toprule
    \textbf{Batch Size}  & \textbf{MSE$\downarrow$} & \textbf{SSIM$\uparrow$} & \textbf{PSNR$\uparrow$}    & \textbf{LPIPS$\downarrow$} \\
    \midrule
      1 & 0.0028 & 0.9999 & 25.5877 & 2.2213e-05 \\
      2 & 0.0037 & 0.9999 & 24.2776 & 5.4648e-06 \\
      4 & 0.0038 & 0.9999 & 24.1785 & 4.4083e-06 \\
      8 & 0.0052 & 0.9998 & 22.7917 & 8.1006e-06 \\
    \bottomrule
    \end{tabular}
\end{table}

\begin{figure*}[!htp]
  \centering
  \includegraphics[scale=0.12]{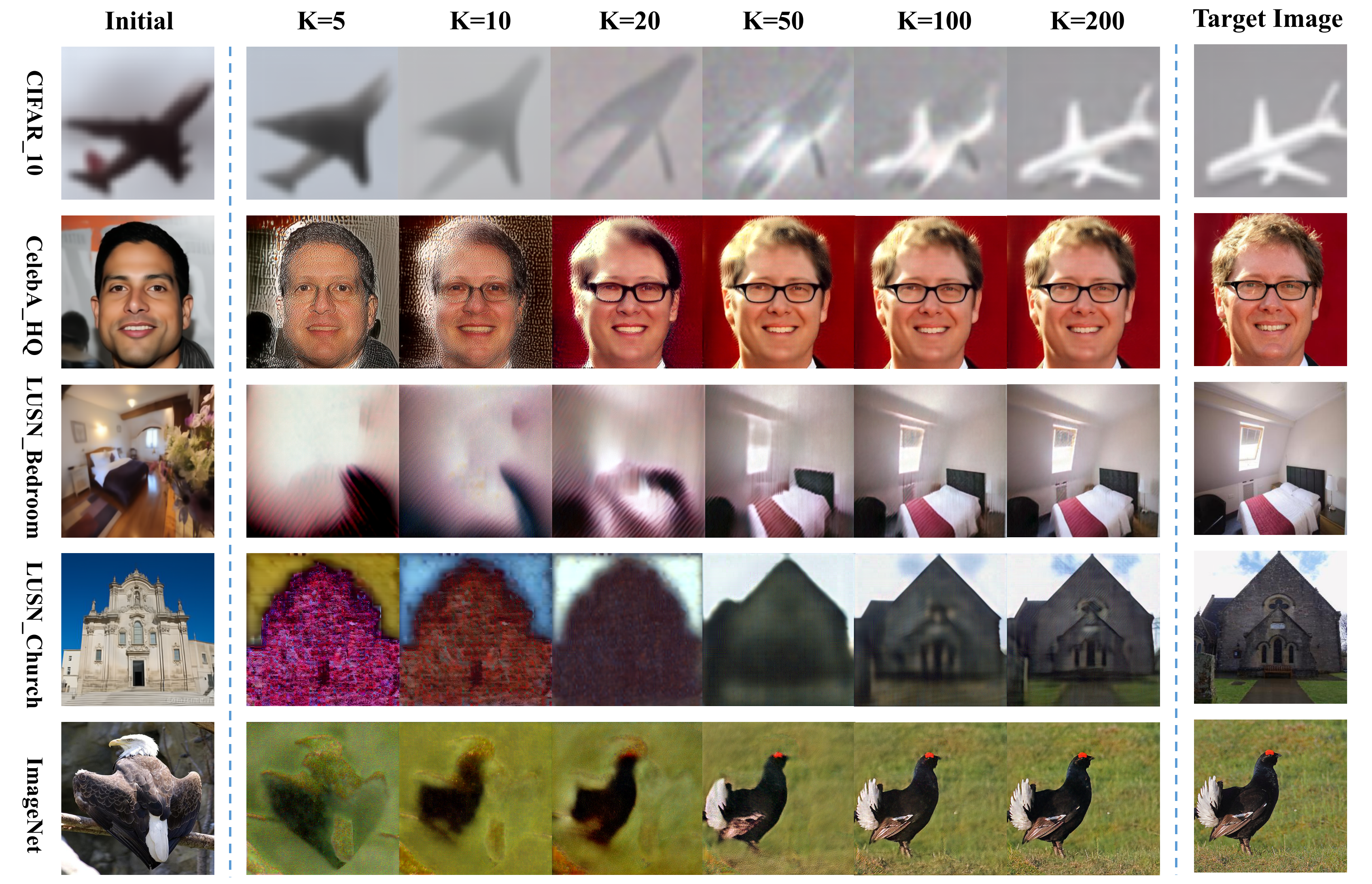}
  \caption{Dependency of reconstruction results on the epochs K.}
  \label{epoch}
\end{figure*}

\begin{figure*}[!htp]
  \centering
  \includegraphics[scale=0.3]{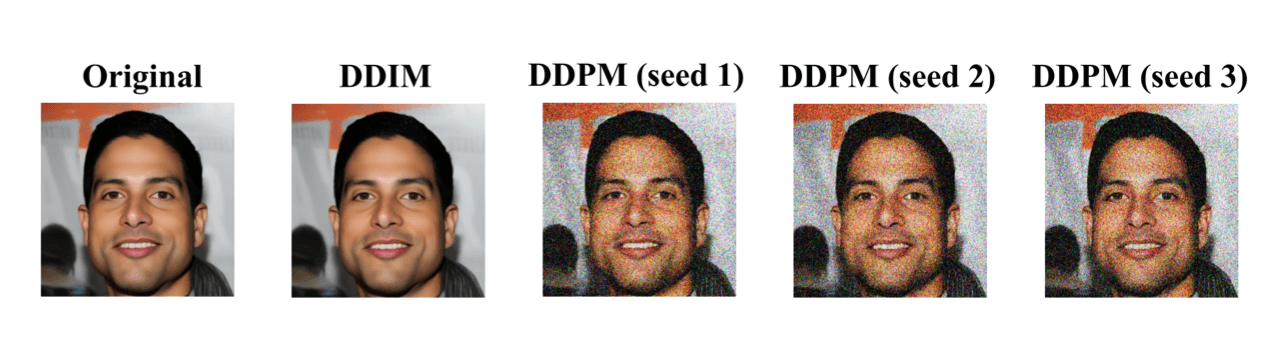}
  \caption{Effect of stochastic manipulation with random seeds.}
  \label{ddpm}
\end{figure*}

\subsubsection{Fine-tuning Epochs K} 
In the fine-tuning process of the gradient-guided diffusion model, we use the Adam optimizer with initial learning rates and decay rate consistent with the default settings in the paper for different datasets. As illustrated in the example of changes across our five datasets in Figure \ref{epoch}, it is evident that, all images generated from the fine-tuned models progressively converge to the ground truth as the epoch K increases.

\subsubsection{Stochastic Manipulations} 
We analyze the impact of using deterministic DDIM as opposed to stochastic DDPM on the original images input into the diffusion model and the images generated by the diffusion model. After inputting a sampled original image into the diffusion model and undergoing forward and reverse processes with distinct sampling methods, the generated images are presented in Figure \ref{ddpm}. We can observe that DDIM sampling produced images that appeared almost identical to the original image. Nevertheless, DDPM, with varying random seeds, leads to images that can be modified in diverse ways, offering potential applications in artistic transfers.

\section{Discussion and Future Work}

In this paper, we introduce a novel reconstruction attack leveraging the advanced image-generative capability of diffusion models. Our proposed gradient-guided diffusion model markedly surpasses the performance of existing state-of-the-art (SOTA) attack methodologies. We conceptualize the target model in this context as an encoder, compressing original images into pertinent gradient 'embeddings'. Conversely, the reconstruction attack method functions as a decoder, extracting images from these gradient 'embeddings'. Looking ahead, we aim to study the adversarial principles between differential privacy protection and reconstruction attacks based on diffusion models. In addition, we will develop efficient strategies for reconstructing $1024 \times 1024$ pixels or even higher resolution images, we plan to replace the U-net in diffusion model with other powerful architectures like Vision Transformer.






%

\bibliographystyle{IEEEtran}
\bibliography{sp.bib}

\appendices

\section{Hyperparameter and Ablation Study}\label{appendix_ablation}
\subsection{Dependency on $S_{\text{for}}$, $S_{\text{gen}}$ and $t_{\text{0}}$}

In Table~\ref{table:2}, the ability of DDIM to reconstruct input images from the latents through the reverse process is evaluated using MSE, SSIM, PSNR, and LPIPS on CelebA-HQ. The ground truth is the original image input to the diffusion model. As $S_{\text{for}}$ and $S_{\text{gen}}$ increase, the reconstruction quality increases. However, when $S_{\text{for}} < S_{\text{gen}}$, the reconstruction quality either remains at a similar level or even experiences a decline. For a fixed pair of values $(S_{\text{for}}, S_{\text{gen}})$, as the return step $t_0$ increases, the reconstruction quality diminishes. This decline can be attributed to the expanding intervals between the adjacent steps, resulting in a notable decrease in quality.

\begin{table*}[!htb]
\caption{Quantitative analysis on reconstruction quality of diffusion model with respect to $S_{\text{for}}$, $S_{\text{gen}}$ and $t_{\text{0}}$.}
\label{table:2}
\centering
\begin{tabular}{c|c|c|cccc}
\hline
\toprule
\textbf{Return Steps}       & \textbf{Forward Steps}    & \textbf{Reverse Steps} & \multicolumn{4}{c}{\textbf{Evaluation Metrics}} \\ \cline{4-7} 
\textbf{$t_{\text{0}}$}       & \textbf{$S_{\text{for}}$}    & \textbf{$S_{\text{gen}}$} & \textbf{MSE$\downarrow$} & \textbf{SSIM$\uparrow$} & \textbf{PSNR$\uparrow$}    & \textbf{LPIPS$\downarrow$} \\
\midrule
\multirow{9}{*}{200}       & \multirow{3}{*}{6}   & 2  & 0.0007  & 0.99998     & 31.3604 & 2.3050e-06           \\    
                           &                      & 6  & 0.0002  & 0.99999     & 36.5098 & 3.3773e-06
                           \\
                           &                      & 10 & 0.0009  & 0.99998     & 30.4322 & 2.5734e-06           \\ 
                           \cline{2-7} 
                           & \multirow{3}{*}{40}  & 2  & 0.0005  & 0.99998     & 32.7429 & 8.1442e-06           \\
                           &                      & 6  & 0.0002  & 0.99999     & 36.1903 & 3.0977e-06           \\
                           &                      & 10 & 0.0002  & 0.99999     & 36.3418 & 2.5328e-06           \\
                           \cline{2-7} 
                           & \multirow{3}{*}{200} & 2  & 0.0005  & 0.99998     & 33.0115 & 7.8355e-06           \\
                           &                      & 6  & 0.0002  & 0.99999     & 36.7999 & 2.6616e-06           \\
                           &                      & 10 & 0.0002  & 1.00000     & 37.0087 & 2.0854e-06           \\
                           \cline{1-7} 
\multirow{9}{*}{300}       & \multirow{3}{*}{6}   & 2  & 0.0015  & 0.99993     & 28.2697 & 1.0684e-05          \\
                           &                      & 6  & 0.0011  & 0.99997     & 29.6959 & 3.5312e-06          \\
                           &                      & 10 & 0.0014  & 0.99996     & 28.4151 & 4.0628e-06           \\
                           \cline{2-7} 
                           & \multirow{3}{*}{40}  & 2  & 0.0009  & 0.99996     & 30.6985 & 1.0798e-05           \\
                           &                      & 6  & 0.0003  & 0.99999     & 35.0346 & 4.5843e-06           \\
                           &                      & 10 & 0.0003  & 0.99999     & 35.5441 & 3.5147e-06           \\
                           \cline{2-7} 
                           & \multirow{3}{*}{200} & 2  & 0.0008  & 0.99996     & 31.1361 & 1.0441e-05           \\
                           &                      & 6  & 0.0003  & 0.99999     & 35.7713 & 3.8538e-06           \\
                           &                      & 10 & 0.0002  & 0.99999     & 36.4015 & 2.7612e-06           \\
                           \cline{1-7} 
\multirow{9}{*}{400}       & \multirow{3}{*}{6}   & 2  & 0.0028  & 0.99988     & 25.5784 & 1.4036e-05           \\
                           &                      & 6  & 0.0015  & 0.99995     & 28.3447 & 5.0611e-06           \\
                           &                      & 10 & 0.0021  & 0.99994     & 26.8234 & 6.3390e-06           \\
                           \cline{2-7} 
                           & \multirow{3}{*}{40}  & 2  & 0.0014  & 0.99993     & 28.6602 & 1.3074e-05           \\
                           &                      & 6  & 0.0004  & 0.99999     & 33.9478 & 6.0451e-06           \\
                           &                      & 10 & 0.0003  & 0.99999     & 34.8467 & 4.4816e-06           \\
                           \cline{2-7} 
                           & \multirow{3}{*}{200} & 2  & 0.0012  & 0.99994     & 29.3381 & 1.2695e-05           \\
                           &                      & 6  & 0.0003  & 0.99999     & 34.7904 & 5.0352e-06           \\
                           &                      & 10 & 0.0003  & 0.99999     & 35.8560 & 3.4288e-06           \\
                           \cline{1-7} 
\multirow{9}{*}{500}       & \multirow{3}{*}{6}   & 2  & 0.0057  & 0.99975     & 22.4317 & 2.7311e-05           \\
                           &                      & 6  & 0.0021  & 0.99991     & 26.8271 & 8.0599e-06          \\
                           &                      & 10 & 0.0028  & 0.99991     & 25.5294 & 9.5305e-06           \\
                           \cline{2-7} 
                           & \multirow{3}{*}{40}  & 2  & 0.0026  & 0.99987     & 25.9191 & 1.7688e-05           \\
                           &                      & 6  & 0.0006  & 0.99998     & 32.5316 & 7.7284e-06          \\
                           &                      & 10 & 0.0004  & 0.99998     & 33.8875 & 5.5397e-06          \\
                           \cline{2-7}
                           & \multirow{3}{*}{200} & 2  & 0.0021  & 0.99988     & 26.7900 & 1.6674e-05           \\
                           &                      & 6  & 0.0005  & 0.99998     & 33.4672 & 6.4426e-06           \\
                           &                      & 10 & 0.0003  & 0.99999     & 35.0376 & 4.2234e-06           \\
\bottomrule
\end{tabular}
\end{table*}

In addition, in Table~\ref{table:1}, the reconstruction of the private images through our gradient-guided fine-tuned diffusion model are also evaluated on CelebA-HQ. The current ground truth is the target image of our attack. The evaluation metrics show the similarity between the results of our attack at iteration 200 and the ground truth. We can observe that, there is no definite correlation between the performance of the results and the value of $(S_{\text{for}}, S_{\text{gen}})$. But a larger value of $t_0$ corresponds to better performance when $(S_{\text{for}}, S_{\text{gen}})$ attains the optimal value at the corresponding $t_0$. This may be attributed to the fact that a smaller value of $t_0$ benefits the reconstruction from the latent diffusion model as mentioned above. Therefore, when $t_0$ is small, the diffusion model retains more information about the images input to the diffusion model, then the images generated by the diffusion model tend to be more similar to the input images rather than the target images of our attack. So, for our purpose of reconstructing the target image, the larger the $t_0$, the less similarity there is between the image generated by the diffusion model and the sampled image inputted to the diffusion model, making it more advantageous for reconstructing the target image. As we can see, when $t_0$ is set to 500, which is the same as our experimental configuration, the optimal configuration is $(S_{\text{for}}, S_{\text{gen}})=(40, 6)$, which is what we keep the same within our settings.

\begin{table*}[!htb]
\caption{Quantitative analysis on the quality of attack reconstruction of the gradient-guided diffusion model with respect to $S_{\text{for}}$, $S_{\text{gen}}$ and $t_{\text{0}}$ of the diffusion model.}
\label{table:1}
\centering
\begin{tabular}{c|c|c|cccc}
\hline
\toprule
\textbf{Return Steps}       & \textbf{Forward Steps}    & \textbf{Reverse Steps} & \multicolumn{4}{c}{\textbf{Evaluation Metrics}} \\ \cline{4-7} 
\textbf{$t_{\text{0}}$}       & \textbf{$S_{\text{for}}$}    & \textbf{$S_{\text{gen}}$} & \textbf{MSE$\downarrow$} & \textbf{SSIM$\uparrow$} & \textbf{PSNR$\uparrow$}    & \textbf{LPIPS$\downarrow$} \\
\midrule
\multirow{9}{*}{200}       & \multirow{3}{*}{6}   & 2  & 0.0102  & 0.99900     & 19.8972 & 5.8712e-05           \\
                           &                      & 6  & 0.0210  & 0.99786     & 16.7852 & 1.3617e-04           \\
                           &                      & 10 & 0.0087  & 0.99910     & 20.6273 & 4.7529e-05           \\
                           \cline{2-7} 
                           & \multirow{3}{*}{40}  & 2  & 0.0189  & 0.99812     & 17.2418 & 1.0779e-04           \\
                           &                      & 6  & 0.0091  & 0.99905     & 20.3992 & 5.0983e-05           \\
                           &                      & 10 & 0.0169  & 0.99823     & 17.7311 & 9.1358e-05           \\
                           \cline{2-7} 
                           & \multirow{3}{*}{200} & 2  & 0.0198  & 0.99803     & 17.0306 & 1.1680e-04           \\
                           &                      & 6  & 0.0166  & 0.99828     & 17.8005 & 9.0135e-05           \\
                           &                      & 10 & 0.0090  & 0.99906     & 20.4578 & 4.9264e-05           \\
                           \cline{1-7} 
\multirow{9}{*}{300}       & \multirow{3}{*}{6}   & 2  & 0.0051  & 0.99966     & 22.9124 & 3.9289e-05          \\
                           &                      & 6  & 0.0036  & 0.99976     & 24.4080 & 2.4979e-05          \\
                           &                      & 10 & 0.0036  & 0.99976     & 24.4015 & 2.3633e-05           \\
                           \cline{2-7} 
                           & \multirow{3}{*}{40}  & 2  & 0.0107  & 0.99909     & 19.7216 & 7.9154e-05           \\
                           &                      & 6  & 0.0098  & 0.99914     & 20.0792 & 6.6664e-05           \\
                           &                      & 10 & 0.0086  & 0.99926     & 20.6649 & 5.0963e-05           \\
                           \cline{2-7} 
                           & \multirow{3}{*}{200} & 2  & 0.0107  & 0.99909     & 19.7234 & 7.9175e-05           \\
                           &                      & 6  & 0.0090  & 0.99920     & 20.4642 & 5.6060e-05           \\
                           &                      & 10 & 0.0036  & 0.99976     & 24.3789 & 2.6513e-05           \\
                           \cline{1-7} 
\multirow{9}{*}{400}       & \multirow{3}{*}{6}   & 2  & 0.0080  & 0.99938     & 20.9761 & 8.0830e-05           \\
                           &                      & 6  & 0.0065  & 0.99956     & 21.8678 & 4.9451e-05           \\
                           &                      & 10 & 0.0027  & 0.99988     & 25.6400 & 2.1328e-05           \\
                           \cline{2-7} 
                           & \multirow{3}{*}{40}  & 2  & 0.0079  & 0.99941     & 21.0512 & 7.3667e-05           \\
                           &                      & 6  & 0.0068  & 0.99952     & 21.6865 & 5.1146e-05           \\
                           &                      & 10 & 0.0117  & 0.99906     & 19.3146 & 8.2489e-05           \\
                           \cline{2-7} 
                           & \multirow{3}{*}{200} & 2  & 0.0077  & 0.99943     & 21.1197 & 7.3990e-05           \\
                           &                      & 6  & 0.0029  & 0.99987     & 25.3701 & 2.3766e-05           \\
                           &                      & 10 & 0.0061  & 0.99960     & 22.1499 & 3.9476e-05           \\
                           \cline{1-7} 
\multirow{9}{*}{500}       & \multirow{3}{*}{6}   & 2  & 0.0064  & 0.99963     & 21.9209 & 6.2177e-05           \\
                           &                      & 6  & 0.0067  & 0.99957     & 21.7382 & 4.1862e-05          \\
                           &                      & 10 & 0.0031  & 0.99988     & 25.0538 & 2.3401e-05           \\
                           \cline{2-7} 
                           & \multirow{3}{*}{40}  & 2  & 0.0089  & 0.99936     & 20.5139 & 6.8990e-05           \\
                           &                      & 6  & 0.0027  & 0.99989     & 25.6804 & 2.2653e-05          \\
                           &                      & 10 & 0.0028  & 0.99989     & 25.4935 & 2.3956e-05          \\
                           \cline{2-7}
                           & \multirow{3}{*}{200} & 2  & 0.0042  & 0.99979     & 23.7244 & 3.6816e-05           \\
                           &                      & 6  & 0.0061  & 0.99963     & 22.1448 & 3.9523e-05           \\
                           &                      & 10 & 0.0028  & 0.99988     & 25.4759 & 2.5137e-05           \\
\bottomrule
\end{tabular}
\end{table*}

\section{Running Time and Resources}\label{appendix.d}
Here, we present details on the fine-tuning running time using the NVIDIA RTX A6000 and usage of GPU resources when handling images of different sizes ($32 \times 32$, $256 \times 256$, and $512 \times 512$ pixels). We utilize CIFAR\_10, CelebA-HQ, and ImageNet, respectively corresponding to images with resolution of $32 \times 32$, $256 \times 256$, and $512 \times 512$ pixels.
\subsection{Running Time}\label{appendix.d.1.}
\subsubsection{Gradient-Guided Diffusion Model Fine-tuning} 
The fine-tuning process of the gradient-guided diffusion model consists of two main procedures: latents precomputing and model fine-tuning. The latents precomputing procedure is performed only once for the same pre-trained diffusion model. With our experimental setup using $S_{\text{for}}$ of 40, on average, the time to precompute latent for each image in CIFAR\_10, CelebA-HQ, and ImageNet is 1.0482, 2.7151, and 8.5559 seconds, respectively. As for the model fine-tuning process, each fine-tuning iteration, which includes the reverse generative process, obtaining of gradients corresponding to generated images, loss calculation, and update of diffusion model parameters. In CIFAR\_10 and CelebA-HQ, the batch size is 1, and $S_{\text{gen}}$ is 6. In ImageNet, the batch size is 1, and $S_{\text{gen}}$ is 2. The process of fine-tuning takes 0.3702, 4.3527, and 25.0031 seconds in CIFAR\_10, CelebA-HQ and ImageNet, respectively. Therefore, when the total epoch K equals 200, the total time taken is 74.0482, 870.5473, and 5000.6248 seconds, respectively. We observe that the size of the image significantly impacts the fine-tuning. In the case of ImageNet, the total time for 200 epochs exceeds one hour, despite having $S_{\text{gen}}$ set to only 2.

\subsubsection{Image Reconstruction via Fine-tuned Diffusion Model} 

In the manipulation of both $32 \times 32$ and $256 \times 256$ images, with parameter settings $(S_{\text{for}},S_{\text{gen}})=(40,6)$, the forward process consumes 1.0482 seconds and 2.7151 seconds, and the generative process takes 0.0767 seconds and 0.2992 seconds, resulting in total times of 1.1249 seconds and 3.0143 seconds, respectively. For $512 \times 512$ images, with $(S_{\text{for}},S_{\text{gen}})=(40,2)$, the forward process and the generative process individually take 8.5559 seconds and 1.0148 seconds, resulting in a total time of 9.7147 seconds.

\subsection{Usage of GPU Resources}

\begin{figure}[!htbp]
  \centering
  \includegraphics[scale=0.26]{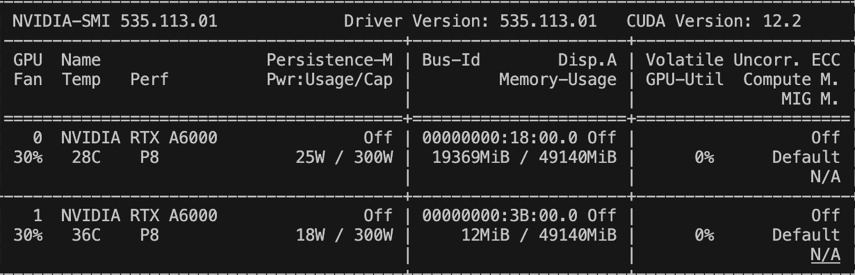}
  \caption{GPU utilization before conducting the experiment. We conduct the experiment on GPU 1: NVIDIA 36C.}
  \label{gpu_before}
\end{figure}

All experiments are conducted on a GPU machine equipped with 4 NVIDIA RTX A6000 GPUs. Prior to initiating our experiments, an overview of GPU usage is presented in Figure \ref{gpu_before}. The figure illustrates that GPUs 1 to 3 exhibit spare capacity, with only 12 MiB out of 49140 MiB being occupied on each. Consequently, our experiments are executed on GPU 1: NVIDIA 36C.

\begin{figure}[!htbp]
  \centering
  \includegraphics[scale=0.26]{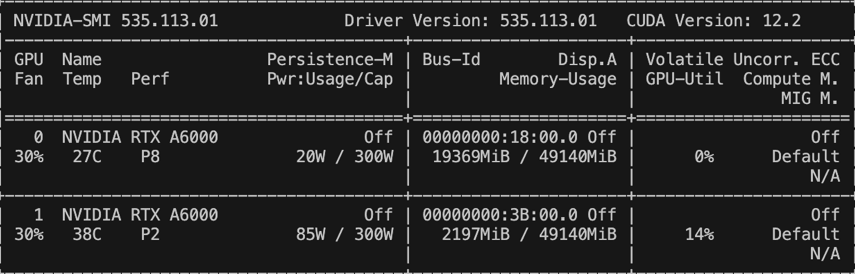}
  \caption{GPU maximum occupancy during the experiment on the dataset with $32 \times 32$ pixel images (CIFAR\_10). We conduct the experiment on GPU 1: NVIDIA 36C.}
  \label{gpu_cifar}
\end{figure}

\begin{figure}[htb]
  \centering
  \includegraphics[scale=0.26]{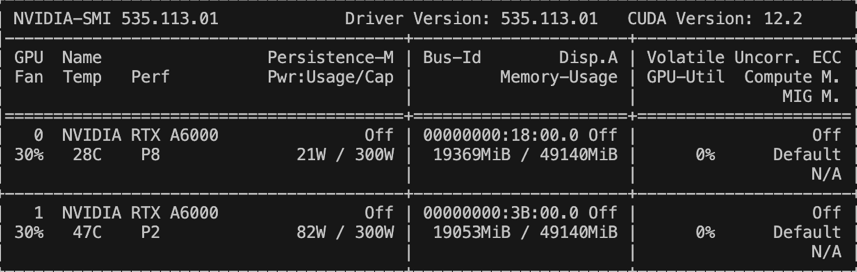}
  \caption{GPU maximum occupancy during the experiment on the dataset with $256 \times 256$ pixel images (CelebA-HQ). We conduct the experiment on GPU 1: NVIDIA 36C.}
  \label{gpu_celeba}
\end{figure}

\begin{figure}[htb]
  \centering
  \includegraphics[scale=0.26]{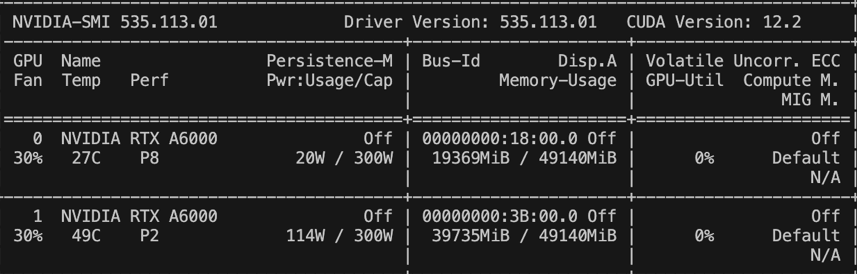}
  \caption{GPU maximum occupancy during the experiment on the dataset with $512 \times 512$ pixel images (ImageNet). We conduct the experiment on GPU 1: NVIDIA 36C.}
  \label{gpu_imagenet_2}
\end{figure}

When processing images of different sizes, including $32 \times 32$ pixel images from CIFAR\_10, $256 \times 256$ pixel images from CelebA-HQ, and $512 \times 512$ pixel images from ImageNet, the GPU utilization differs based on the resolution of images. The basic settings remain consistent with the normal configuration outlined in the paper. Specifically, when conducting experiments on $32 \times 32$ pixel images (as illustrated in Figure \ref{gpu_cifar}), the maximum GPU occupancy reaches 2197 MiB out of 49140 MiB. Figure \ref{gpu_celeba} demonstrates that the maximum GPU occupancy is 19053 MiB out of 49140 MiB when conducting experiments on $256 \times 256$ pixel images. Additionally, in Figure \ref{gpu_imagenet_2}, we can find that the maximum GPU occupancy reaches 39735 MiB out of 49140 MiB during experiments on $512 \times 512$ pixel images.

\begin{figure}[htb]
  \centering
  \includegraphics[scale=0.26]{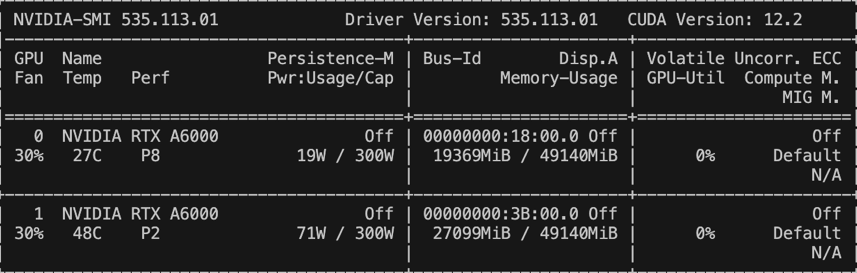}
  \caption{GPU maximum occupancy during the experiment on the dataset with $512 \times 512$ images (ImageNet) when $S_{\text{gen}} = 1$. We conduct the experiment on GPU 1: NVIDIA 36C.}
  \label{gpu_imagenet_1}
\end{figure}

In the datasets with image pixels of $32 \times 32$ and $256 \times 256$, we set $(S_{\text{for}}, S_{\text{gen}})$ to $(40, 6)$. However, for the $512 \times 512$ dataset, $(S_{\text{for}}, S_{\text{gen}})$ is configured as $(40, 2)$. These parameter settings are chosen not only for the effectiveness of reconstruction results but also due to constraint imposed by the GPU's DRAM. Larger values for $S_{\text{gen}}$ are not feasible within this limitation. As previously illustrated, with the parameter settings $(S_{\text{for}}, S_{\text{gen}}) = (40, 2)$ in ImageNet, the maximum GPU occupancy reaches 39735 MiB out of 49140 MiB. To further explore DRAM usage during each step of the reverse process, we conduct an additional experiment by setting $S_{\text{gen}}$ to 1, then the GPU occupancy differs from the case when $S_{\text{gen}} = 2$ only by the DRAM required for one step of the reverse process of the diffusion model. The GPU usage in this case, as depicted in Figure \ref{gpu_imagenet_1}, reveals a maximum GPU occupancy of 27099 MiB out of 49140 MiB. Therefore, one step in the reverse process requires approximately 12633 MiB of VRAM. Due to limitations in the GPU's VRAM, assigning a larger value to $S_{\text{gen}}$ is not feasible. Consequently, our future research will focus on developing efficient strategies and methods to optimize GPU efficiency.

\end{document}